\RequirePackage{fix-cm}
\documentclass[twocolumn]{svjour3}          
\smartqed  
\usepackage{natbib}
\usepackage{graphicx}
\usepackage{hyperref} 
\usepackage[utf8]{inputenc} 
\usepackage[T1]{fontenc}    
\usepackage{url}            
\usepackage{booktabs}       
\usepackage{amsfonts}       
\usepackage{nicefrac}       
\usepackage{microtype}      
\usepackage{epsfig}
\usepackage{float}
\usepackage{multicol}
\usepackage{multirow}
\usepackage{amssymb}
\usepackage{amsmath}
\usepackage{wrapfig}
\usepackage[toc,page]{appendix}
\usepackage{xspace}
\usepackage{color}
\usepackage[dvipsnames]{xcolor}
\usepackage[misc]{ifsym}
\usepackage{makecell}

\newcommand{\ie}{i.e}
\newcommand{\eg}{e.g}

\newcommand{\alambic}{\includegraphics[width=0.02\linewidth]{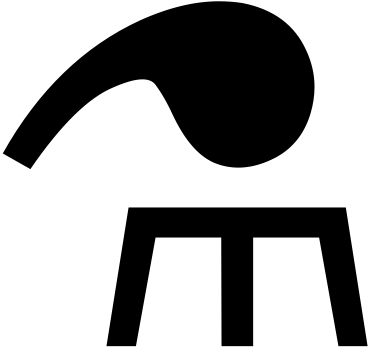}\xspace}

\def\onedot{.\xspace}
\def\eg{\emph{e.g}\onedot} 

\def\ie{\emph{i.e}\onedot}

\def\etc{\emph{etc}\onedot}

\begin{document}

\title{ViTAEv2: Vision Transformer Advanced by Exploring Inductive Bias for Image Recognition and Beyond

\thanks{*Q. Zhang and Y. Xu  contribute equally to this work.\\Mr. Qiming Zhang, Mr Yufei Xu and Dr Jing Zhang are supported by the Australian Research Council Research Project FL-170100117.}
}

\author{Qiming Zhang$^{1*}$         \and
        Yufei Xu$^{1*}$ \and
        Jing Zhang$^{1}$ \and
        Dacheng Tao$^{2,1}$
}

\authorrunning{Qiming Zhang, et al.}
\institute{
Qiming Zhang (qzha2506@uni.sydney.edu.au) \\
Yufei Xu (yuxu71166@uni.sydney.edu.au) \\
Jing Zhang (jing.zhang1@sydney.edu.au) \\
\Letter~Dacheng Tao (dacheng.tao@gmail.com)\\
$^{1}$ School of Computer Science,
Faculty of Engineering, The University of Sydney, Darlington, NSW 2008, Australia.\\
$^{2}$ JD Explore Academy, China \\
}

\date{Received: date / Accepted: date}
\maketitle

\begin{abstract}
Vision transformers have shown great potential in various computer vision tasks owing to their strong capability to model long-range dependency using the self-attention mechanism. Nevertheless, they treat an image as a 1D sequence of visual tokens, lacking an intrinsic inductive bias (IB) in modeling local visual structures and dealing with scale variance, which is instead learned implicitly from large-scale training data with longer training schedules. In this paper, we leverage the two IBs and propose the ViTAE transformer, which utilizes a reduction cell for multi-scale feature and a normal cell for locality. The two kinds of cells are stacked in both isotropic and multi-stage manners to formulate two families of ViTAE models, i.e., the vanilla ViTAE and ViTAEv2. Experiments on the ImageNet dataset as well as downstream tasks on the MS COCO, ADE20K, and AP10K datasets validate the superiority of our models over the baseline and representative models. Besides, we scale up our ViTAE model to 644M parameters and obtain the state-of-the-art classification performance, i.e., 88.5\% Top-1 classification accuracy on ImageNet validation set and the best 91.2\% Top-1 classification accuracy on ImageNet Real validation set, without using extra private data. It demonstrates that the introduced inductive bias still helps when the model size becomes large. The source code and pretrained models are publicly available at \href{https://github.com/ViTAE-Transformer/ViTAE-Transformer}{code}.

\end{abstract}

\section{Introduction}
\label{intro}

Transformers~\citep{vaswani2017attention,devlin2018bert} have become the popular frameworks in NLP studies owing to their strong ability in modeling long-range dependencies by the self-attention mechanism. Such success and good properties of transformers have inspired many following works that apply them in various computer vision tasks~\citep{dosovitskiy2020image,zheng2020rethinking,wang2021pyramid}. Among them, ViT~\citep{dosovitskiy2020image} is the pioneering work that adapts a pure transformer model for vision by embedding images into a sequence of visual tokens and modeling the global dependencies among them with stacked transformer blocks. Although it achieves promising performance on image classification, it experiences a severe data-hungry issue, \ie, requiring large-scale training data and a longer training schedule for better performance. One important reason is that ViT does not efficiently utilize the prior knowledge in vision tasks and lacks such inductive bias (IB) in modeling local visual clues (\eg, edges and corners) and dealing with objects at various scales like convolutions. Alternatively, ViT has to learn such IB implicitly from large-scale data.

\begin{figure}
    \includegraphics[width=\linewidth]{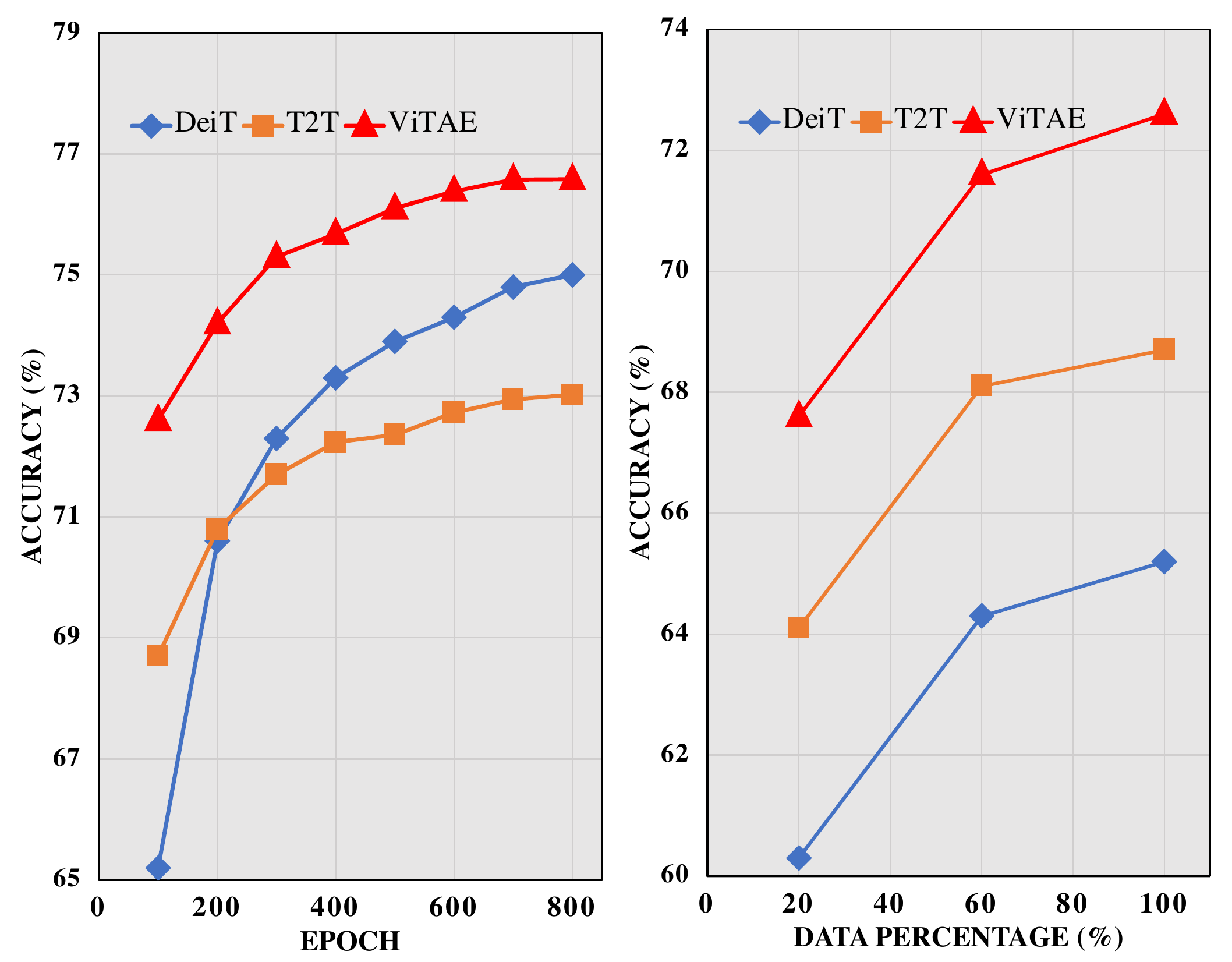}
    \caption{Comparison of data and training efficiency of DeiT-T~\citep{touvron2020training}, T2T-ViT-7~\citep{yuan2021tokens} and ViTAE-T on ImageNet.}
    \label{fig:efficiency}
\end{figure}

Unlike vision transformers, Convolution Neural Networks (CNNs) are naturally equipped with the intrinsic IBs of locality and scale-invariance and still serve as prevalent backbones in vision tasks~\citep{he2016deep,szegedy2017inception,chen2017rethinking,zhao2017pyramid}. The success of CNNs inspires us to explore the benefits of introducing intrinsic IBs in vision transformers. We start by analyzing the above two IBs of CNNs, \ie, locality and scale-invariance. Convolution that computes local correlation among neighbor pixels is good at extracting local features such as edges and corners. Consequently, CNNs can provide great low-level features at the shallow layers \citep{zeiler2014visualizing}, which are then aggregated into high-level features progressively by a bulk of sequential convolutions \citep{huang2017densely,simonyan2014very,szegedy2015going}. Moreover, CNNs have a hierarchy structure to extract multi-scale features at different layers \citep{simonyan2014very,krizhevsky2012imagenet,he2016deep}. Intra-layer convolutions can also learn features at different scales by varying their kernel sizes and dilation rates~\citep{he2015spatial,szegedy2017inception,chen2017rethinking,lin2017feature,zhao2017pyramid}. Consequently, scale-invariant feature representations can be obtained via intra- or inter-layer feature fusion. Nevertheless, CNNs are not well suited to model long-range dependencies\footnote{Despite the projection layer in a transformer can be viewed as $1\times 1$ convolution~\citep{chen2021empirical}, the term of convolution here refers to those with larger kernels, \eg, $3 \times 3$, which are widely used in typical CNNs to extract spatial features.}, which is the key advantage of transformers. An interesting question then comes up: can we improve vision transformers by leveraging the good properties of CNNs? Recently, DeiT~\citep{touvron2020training} explores the idea of distilling knowledge from CNNs to transformers to facilitate training and improve performance. However, it requires an off-the-shelf CNN model as the teacher and incurs extra training costs.

Different from DeiT, we explicitly introduce intrinsic IBs into vision transformers by re-designing the network structures in this paper. Current vision transformers always obtain tokens with single-scale context~\citep{dosovitskiy2020image,yuan2021tokens,wang2021pyramid,liu2021swin} and learn to adapt to objects at different scales from data. For example, T2T-ViT \citep{yuan2021tokens} improves ViT by delicately generating tokens in a soft split manner. Specifically, it uses a series of Tokens-to-Token transformation layers to aggregate single-scale neighboring contextual information and progressively structures the image to tokens.
Motivated by the success of CNNs in dealing with scale variance, we explore a similar design in transformers, \ie, intra-layer convolutions with different receptive fields \citep{szegedy2017inception,yu2017dilated}, to embed multi-scale context into tokens. Such a design allows tokens to carry useful features of objects at various scales, thereby naturally having the intrinsic scale-invariance IB and explicitly facilitating transformers to learn scale-invariant features more efficiently from data. On the other hand, low-level local features are fundamental elements to generate high-level discriminative features. Although transformers can also learn such features at shallow layers from data, they are not skilled as convolutions by design. Recently, \citep{yan2021contnet,li2021localvit,graham2021levit} stack convolutions and attention layers sequentially and demonstrate that locality is a reasonable compensation of global dependency. However, this serial structure ignores the global context during locality modeling (and vice versa). To avoid such a dilemma, we follow the ``divide-and-conquer'' idea and propose modeling locality and long-range dependencies in parallel and fusing the features to account for both. In this way, we empower transformers to learn local and long-range features within each block more effectively. 

Technically, we propose a new \textbf{Vi}sion \textbf{T}ransformers \textbf{A}dvanced by \textbf{E}xploring Intrinsic Inductive Bias (\textit{\textbf{ViTAE}}), which is a combination of two types of basic cells, \ie, reduction cell (RC) and normal cell (NC). RCs are used to downsample and embed the input images into tokens with rich multi-scale context, while NCs aim to jointly model locality and global dependencies in the token sequence. Moreover, these two types of cells share a simple basic structure, \ie, paralleled attention module and convolutional layers followed by a feed-forward network (FFN). It is noteworthy that RC has an extra pyramid reduction module with atrous convolutions of different dilation rates to embed multi-scale context into tokens. Following the setting in~\citep{yuan2021tokens}, we stack three reduction cells to reduce the spatial resolution by $1/16$ and a series of NCs to learn discriminative features from data. ViTAE outperforms representative vision transformers in terms of data efficiency and training efficiency (see Figure~\ref{fig:efficiency}) as well as classification accuracy and generalization on downstream image classification tasks. In addition, we further scale up ViTAE to large models and show that the inductive bias still helps to obtain better performance, e.g., ViTAE-H with 644M parameters achieves 88.5\% Top-1 classification accuracy on ImageNet without using extra private data.

Beyond image classification, backbone networks should adapt well to various downstream tasks such as object detection, semantic segmentation, and pose estimation. To this end, we extend the vanilla ViTAE to the multi-stage design, \ie, ViTAEv2. Specifically, a natural choice is to construct the model by re-arranging the reduction cells and normal cells according to the strategies in \citep{wang2021pyramid,liu2021swin} to have multi-scale feature outputs, \ie, several consecutive NC cells are used following one RC module at each stage (feature resolution) rather than using a series of NCs only at the last stage. As a result, the multi-scale features from different stages can be utilized for those various downstream tasks. One remaining issue is that the vanilla attention operations in transformers have a quadratic computational complexity, requiring a large memory footprint and computation cost, especially for feature maps with a large resolution. To mitigate this issue, we further explore another inductive bias, \ie, local window attention introduced in~\citep{liu2021swin}, in the RC and NC modules. Since the parallel convolution branch in the proposed two cells can encode position information and enable inter-window information exchange, special designs like the relative position encoding and window-shifting mechanism in~\citep{liu2021swin} can be omitted. Consequently, our ViTAEv2 models outperform state-of-the-art methods for various vision tasks, including image classification, object detection, semantic segmentation, and pose estimation, while keeping a fast inference speed and reasonable memory footprint.

The contribution of this study is threefold. \textbf{First}, we explore two types of intrinsic IB in transformers, \ie, scale invariance and locality, and demonstrate the effectiveness of this idea by designing a new transformer architecture named ViTAE based on two new reduction and normal cells that incorporate the above two IBs. ViTAE outperforms representative vision transformers regarding classification accuracy, data efficiency, training efficiency, and generalization on downstream vision tasks. \textbf{Second}, we scale up our ViTAE model to 644M parameters and obtain 88.5\% Top-1 classification accuracy on ImageNet without using extra private data, which is better than the state-of-the-art Swin Transformer, demonstrating that the introduced inductive bias still helps when the model size becomes large. \textbf{Third}, we extend the vanilla ViTAE to the multi-stage design, \ie, ViTAEv2. It learns multi-scale features at different stages efficiently while keeping a fast inference speed and reasonable memory footprint for large-size input images. Experiments on popular benchmarks demonstrate that it outperforms state-of-the-art methods for various downstream vision tasks, including image classification, object detection, semantic segmentation, and pose estimation. 

The following of this paper is organized as follows. Section 2 describes the relevant works to our paper. We then detail the two basic cells, the vanilla ViTAE model, the scaling strategy for ViTAE, as well as the multi-stage design for ViTAEv2 in Section 3. Next, Section 4 presents the extensive experimental results and analysis. Finally, we conclude our paper in Section 5 and discuss the potential applications and future research directions.

\section{Related Work}
\label{related work}
\subsection{CNNs with intrinsic inductive bias}

CNNs~\citep{krizhevsky2012imagenet,zeiler2014visualizing,he2016deep} have explored several inductive biases with specially designed operations and have led to a series of breakthroughs in vision tasks, such as image classification, object detection, and semantic segmentation. For example, following the fact that local pixels are more likely to be correlated in images \citep{lecun1995convolutional}, the convolution operations in CNNs extract features from the neighbor pixels within the receptive field determined by the kernel size~\citep{lecun2015deep}. By stacking convolution operations, CNNs have the inductive bias in modeling locality naturally.

In addition to the locality, another critical inductive bias in visual tasks is scale-invariance, where multi-scale features are needed to represent the objects at different scales effectively~\citep{luo2016understanding,yu2016multi}. 
For example, to effectively learn features of large objects, a large receptive field is needed by either using large convolution kernels~\citep{yu2016multi,yu2017dilated} or a series of convolution layers in deeper architectures~\citep{he2016deep,huang2017densely,simonyan2014very,szegedy2015going}. However, such operations may ignore the features of small objects. 
To construct multi-scale feature representation for objects at different scales effectively, various image pyramid techniques~\citep{chen2017rethinking,adelson1984pyramid,olkkonen1996gaussian,burt1987laplacian,lai2017deep,demirel2010image} have been explored, where features are extracted from a pyramid of images at different resolutions respectively~\citep{lin2016efficient,chen2017rethinking,ng2003sift,rublee2011orb,ke2004pca,bay2006surf}, either in a hand-crafted manner or learned manner. Accordingly, features from the small-scale images mainly encode the large objects, while features from the large-scale images respond more to small objects. Then, features extracted from different resolutions are fused to form the scale-invariant feature, \ie, the inter-layer fusion. Another way to obtain the scale-invariant feature is to extract and aggregate multi-scale context by using multiple convolutions with different receptive fields in a parallel manner, \ie, the intra-layer fusion~\citep{zhao2017pyramid,szegedy2015going,szegedy2017inception,szegedy2016rethinking}. Either the inter-layer or intra-layer fusion empowers the CNNs with the scale-invariance inductive bias. It helps improve their performance in recognizing objects at different scales.

However, it is unclear whether these inductive biases can help the visual transformer to achieve better performance. This paper explores the possibility of introducing two types of inductive biases in the vision transformer, namely locality by introducing convolution in the vision transformer and scale-invariance by encoding a multi-scale contxt into each visual token using multiple convolutions with different dilation rates, following the convention of intra-layer fusion.

\subsection{Vision transformers with inductive bias}

ViT~\citep{dosovitskiy2020image} is the pioneering work that applies a pure transformer to vision tasks and achieves promising results. It treats images as a 1D sequence, embeds them into several tokens, and then processes them by stacked transformer blocks to get the final prediction. However, since ViT simply treats images as 1D sequences and thus lacks inductive bias in modeling local visual structures, it indeed implicitly learns the IB from a large amount of data. Similar phenomena can also be observed in models with fewer inductive biases in their structures~\citep{tolstikhin2021mlp,el2021xcit,he2021gauge}.

To alleviate the data-hungry issue, the following works explicitly introduce inductive bias into vision transformers, \eg, leveraging the IB from CNNs to facilitate the training of vision transformers with less training data or shorter training schedules. For example, DeiT~\citep{touvron2020training} proposes to distill knowledge from pre-trained CNNs to transformers during training via an extra distillation token to imitate the behavior of CNNs. However, it requires an off-the-shelf CNN model as a teacher, introducing extra computation cost. Recently, some works try to introduce the intrinsic IB of CNNs into vision transformers explicitly~\citep{han2021transformer,peng2021conformer,graham2021levit,li2021localvit,d2021convit,yan2021contnet,wu2021cvt,yuan2021incorporating,chen2021crossvit,liu2021swin}. For example, \citep{li2021localvit,graham2021levit,wu2021cvt,dai2021coatnet} stack convolutions and attention layers sequentially, resulting in a serial structure and modeling the locality and global dependency accordingly. \citep{wang2021pyramid} design sequential multi-stage structures while \citep{liu2021swin} apply attention within local windows. However, these serial structures may ignore the global context during locality modeling (and vice versa). \citep{wang2021crossformer} establishes connection across different scales at the cost of heavy computation. To jointly model global and local context, Conformer~\citep{peng2021conformer} and MobileFormer~\citep{chen2021mobile} employ a model-parallel structure, consisting of parallel individual convolution and transformer branches and a complicated bridge connection between the two branches. Different from them, we follow the ``divide-and-conquer'' idea and propose to model locality and global dependencies simultaneously via a parallel structure within each transformer layer. In this way, the convolution and attention modules are designed to complement each other within the transformer block, which is more beneficial for the models to learn better features for both classification and dense prediction tasks.

\subsection{Self supervised learning and model scaling}

As demonstrated in previous studies, scaled-up models are naturally few-shot learners and beneficial to obtain better performance no matter in language, image, or cross-modal domains~\citep{devlin2018bert,zhai2021scaling,radford2021learning}. 
Recently, many efforts have been made to scale up vision models, \eg, BiT~\citep{kolesnikov2020big} and EfficientNet~\citep{tan2019efficientnet} scale up the CNN models to hundreds of millions of parameters by employing wider and deeper networks, and obtain superior performance on many vision tasks. However, they need to train the scaled-up models with a much larger scale of private data, \ie, JFT300M~\citep{kolesnikov2020big}. Similar phenomena can be observed when training the scaled-up vision transformer models for better performance~\citep{dosovitskiy2020image,zhai2021scaling}. 

However, it is not easy to gather such large amounts of labeled data to train the scaled-up models. On the other hand, self-supervised learning can help train scaled-up models using data without labels. For example, CLIP~\citep{radford2021learning} adopts paired text and image data captured from the Internet and exploits the consistency between text and images to train a big transformer model, which obtains good performance on image and text generation tasks. \citep{liu2019roberta} adopt masked language modeling (MLM) as pretext tasks and generate supervisory signals from the input data. Specifically, they take masked sentences with several words overrode with mask and predicted the masked words with the words from the sentence before masking as supervision. In this way, these models do not require additional labels for the training data and achieve superior performance on translation, sentiment analysis, \etc Inspired by the superior performance of MLM tasks in language, masked image modeling (MIM) tasks have been explored in vision tasks recently. For example, BEiT~\citep{beit} tokenizes the images into visual tokens and randomly masks some tokens using a block-wise manner. The vision transformer model must predict the original tokens for those masked tokens. In this way, BEiT obtains superior classification and dense prediction performance using publicly available ImageNet-22K dataset~\citep{deng2009imagenet}. MAE~\citep{he2021masked} simplifies the requirement of tokenizers and simply treats the image pixels as the targets for reconstruction. Using only ImageNet-1K training data, MAE obtains impressive performance. It is under-explored whether the vision transformers with introduced inductive bias can be scaled up, \eg, in a self-supervised setting. Besides, whether inductive bias can still help these scaled-up models achieve better performance remains unclear. In this paper, we make an attempt to answer this question by scaling up the ViTAE model and training it in a self-supervised manner. Experimental results confirm the value of introducing inductive bias in scaled-up vision transformers.

\subsection{Comparison to the conference version}
A preliminary version of this work was presented in \citep{xu2021vitae}. This paper extends the previous study by introducing three major improvements.

\begin{enumerate}
    \item We scale up the ViTAE model to different model sizes, including ViTAE-B, ViTAE-L, and ViTAE-H. With the help of inductive bias, the proposed ViTAE-H model with 644M parameters obtains the state-of-the-art classification performance, i.e., 88.5\% Top-1 classification accuracy on ImageNet validation set and the best 91.2\% Top-1 classification accuracy on ImageNet Real validation set, without using extra private data. It demonstrates that the introduced inductive bias still helps when the model size becomes large. We also show the excellent few-shot learning ability of the scaled-up ViTAE models.
    \item We extend the vanilla ViTAE to the multi-stage design and devise ViTAEv2. The efficiency of the RC and NC modules is also improved by exploring another inductive bias from local window attention. ViTAEv2 outperforms state-of-the-art models for image classification tasks as well as downstream vision tasks, including object detection, semantic segmentation, and pose estimation.
    \item We also present more ablation studies and experiment analysis regarding module design, inference speed, memory footprint, and comparisons with the latest works.
    
\end{enumerate}

\begin{figure*}[ht]
    \centering
    \includegraphics[width=1.0\linewidth]{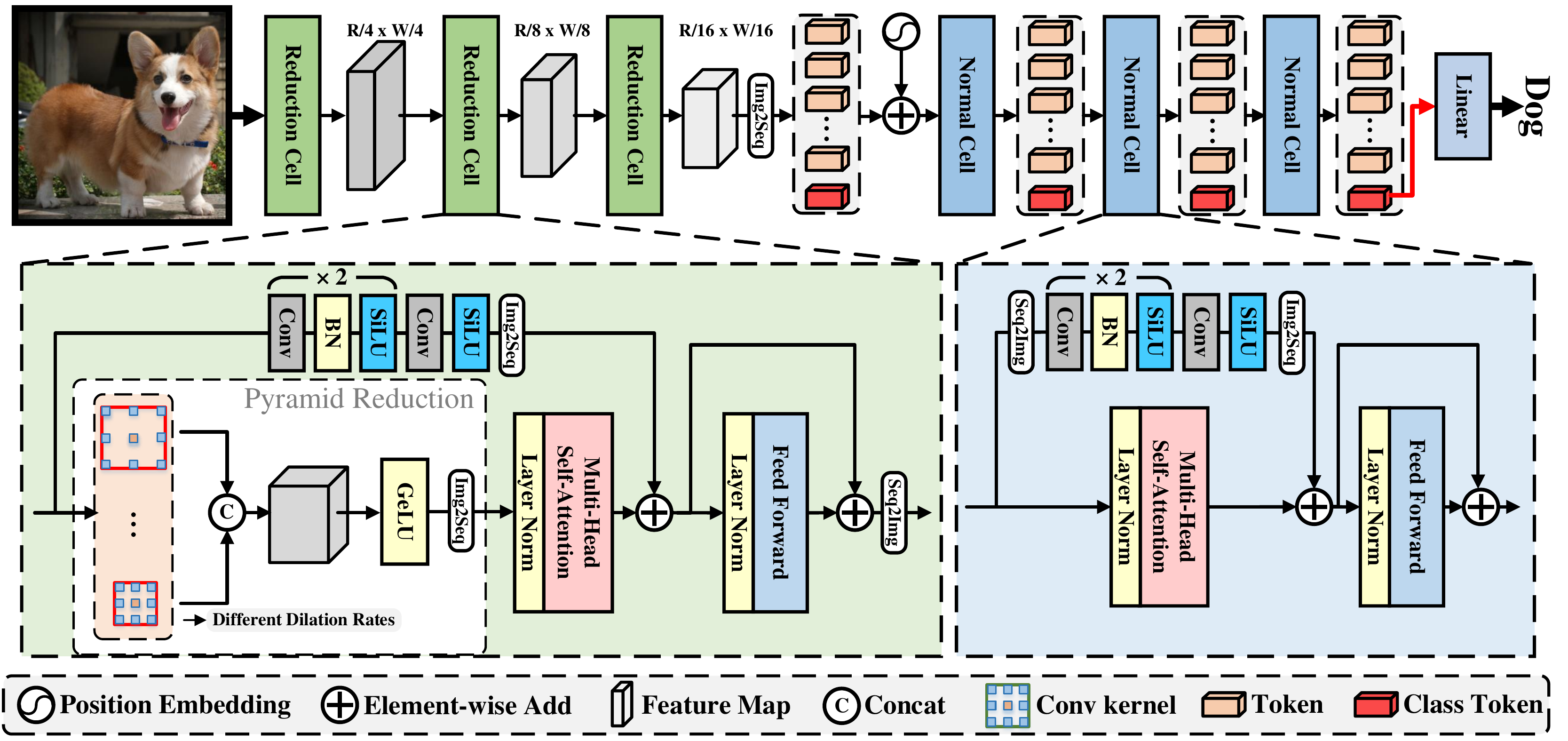}
    \caption{The structure of the proposed ViTAE model in the isotropic manner. It is constructed by stacking three RCs and several NCs. Both types of cells share a simple basic structure, \ie, an MHSA module and a parallel convolutional module followed by an FFN. In particular, RC has an extra pyramid reduction module using atrous convolutions with different dilation rates to embed multi-scale context.}
    \label{fig:structure}
\end{figure*}

\section{Methodology}

\subsection{Revisit vision transformer}
We first give a brief review of the vision transformer in this part. To adapt transformers to vision tasks, ViT~\citep{dosovitskiy2020image} first splits an image $x \in R^{H \times W \times C}$ into several non-overlapping patches with the patch size $p$, and embeds them into visual tokens (\ie, $ x_t \in R^{N \times D} $) in a patch-to-token manner, where $H$, $W$, $C$ denote the height, width, and channel dimensions of the input image respectively, $N$ and $D$ denote the token number and token dimension, respectively, and $N = (H\times W) / p^2$. Then, an extra learnable embedding with the same dimension $D$, considered as a class token, is concatenated to the visual tokens before adding position embeddings to all the tokens in an element-wise manner. In the following part of this paper, we use $x_t$ to represent all tokens, and $N$ is the total number of tokens after concatenation for simplicity unless specified. These tokens are fed into several sequential transformer layers for the final prediction. Each transformer layer is composed of two parts, \ie, a multi-head self-attention module (MHSA) and a feed-forward network (FFN).

MHSA extends single-head self-attention (SHSA) by using different projection matrices for each head. In other words, MHSA is obtained after repeating SHSA for $h$ times, where $h$ is the number of heads. Specifically, for SHSA, the input tokens $x_t$ are first projected to queries ($Q$), keys ($K$) and values ($V$) using three different projection matrices, \ie, $Q,K,V = x_t W_Q, x_t Q_K,$ $ x_t Q_V$, where $W_{Q/K/V} \in R^{D \times \frac{D}{h}}$ denotes the projection matrix for query/key/value, respectively. Then, the self-attention operation is calculated as:
\begin{equation}
    Attention(Q, K, V) = softmax(\frac{QK^T}{\sqrt{D}})V,
\end{equation}
where the output of each head is of size $R^{N \times \frac{D}{h}}$. Then the features of all the $h$ heads are concatenated along the channel dimension and formulate the MHSA module's output.

FFN is placed on top of the MHSA module and applied to each token identically and separately. It consists of two linear transformations with an activation function in between. Besides, a layer normalization~\citep{ba2016layer} and a shortcut are added before and aside from the MHSA and FFN, respectively. 

\subsection{The isotropic design of ViTAE}
ViTAE aims to introduce the intrinsic IB in CNNs to vision transformers. As shown in Figure~\ref{fig:structure}, ViTAE is composed of two types of cells, \ie, RCs and NCs. RCs are responsible for downsampling while embedding multi-scale context and local information into tokens, and NCs are used to further model the locality and long-range dependencies in the tokens. Taken an image $x \in R^{H \times W \times C}$ as input, three RCs are used to gradually downsample $x$ with a total of 16$\times$ ratio by 4$\times$, 2$\times$, and 2$\times$, respectively. Thereby, the output tokens of the RCs after downsampling are of size $[H/16, W/16, D]$ where $D$ is the token dimension (64 in our experiments). The output tokens of RCs are then flattened as $R^{(HW/256) \times D}$, concatenated with the class token (red in the figure), and added by the sinusoid position encoding. Next, the tokens are fed into the following NCs, which keep the length of the tokens. Finally, the prediction probability is obtained using a linear classification layer on the class token from the last NC.

\subsubsection{Reduction cell}
Instead of directly splitting and flattening images into visual tokens based on a linear image patch embedding layer, we devise the reduction cell to embed multi-scale context and local information into visual tokens, introducing the intrinsic scale-invariance and locality IBs from convolutions into ViTAE. Technically, RC has two parallel branches responsible for modeling locality and long-range dependency, followed by an FFN for feature transformation. We denote the input feature of the $i_{th}$ RC as $f_i \in R^{H_i \times W_i \times D_i}$. The input of the first RC is the image $x$. In the global dependency branch, $f_i$ is firstly fed into a Pyramid Reduction Module (PRM) to extract multi-scale context, \ie,
\begin{equation}
\begin{aligned}
    f_i^{ms} &\triangleq PRM_i(f_i) \\ 
    &= Cat([Conv_{ij}(f_i; s_{ij}, r_{i}) | s_{ij} \in \mathcal{S}_i, r_{i} \in \mathcal{R}]),
\end{aligned}
\end{equation}
where $Conv_{ij}(\cdot)$ indicates the $j_{th}$ convolutional layer in the $i_{th}$ PRM (i.e., $PRM_i(\cdot)$). It uses a dilation rate $s_{ij}$ from the predefined dilation rate set $\mathcal{S}_i$ corresponding to the $i$th RC. Note that we use stride convolution to reduce the spatial dimension of features by a ratio $r_{i}$ from the predefined reduction ratio set $\mathcal{R}$. The features after convolution are concatenated along the channel dimension, \ie, $f_i^{ms} \in R^{(W_i/r_i) \times (H_i/r_i) \times (|\mathcal{S}_i|D_i)}$, where $|\mathcal{S}_i|$ denotes the number of dilation rates in the set $\mathcal{S}_i$. $f_i^{ms}$ is then processed by an MHSA module to model long-range dependencies, \ie, 
\begin{equation}
    f_i^{g} = MHSA_i(Img2Seq(f_i^{ms})),
\end{equation}
where $Img2Seq(\cdot)$ is a simple reshape operation to flatten the feature map to a 1D sequence. In this way, $f_i^{g}$ embeds the multi-scale context in each token. Note that the traditional MHSA individually attends each token at the same scale and thus lacks the ability to model the relationship between tokens at different scales. By contrast, the introduced multi-scale convolutions in reduction cells can (1) mitigate the information loss when merging tokens by looking at a larger field and (2) embed multi-scale information into tokens to aid the following MHSA to model the better global dependencies based on features at different scales. 

In addition, we use a Parallel Convolutional Module (PCM) to embed local context within the tokens, which are fused with $f_i^{g}$ as follows:
\begin{equation}
    f_i^{lg} = f_i^{g} + PCM_i(f_i).
\end{equation}
Here, $PCM_i(\cdot)$ represents the PCM of the $i_{th}$ RC, which is composed of an $Img2Seq(\cdot)$ operation and three stacked convolution layers with BN layers and activation layers in between. It is noteworthy that the parallel convolution branch has the same spatial downsampling ratio as the PRM by using stride convolutions. In this way, the token features can carry both local and multi-scale context, implying that RC acquires the locality IB and scale-invariance IB by design. The fused tokens are then processed by the FFN and reshaped back to feature maps, \ie,
\begin{equation}
\label{eq:residual ffn}
    f_{i+1} = Seq2Img(FFN_i(f_i^{lg}) + f_i^{lg}),
\end{equation}
where the $Seq2Img(\cdot)$ is a simple reshape operation to reshape a token sequence back to feature maps. $FFN_i(\cdot)$ represents the FFN in the $i_{th}$ RC. In our ViTAE, three RCs are stacked sequentially to gradually reduce the input image's spatial dimension by 4$\times$, 2$\times$, and 2$\times$, respectively. As the first RC handles images with high resolution, we adopt Performer~\citep{choromanski2020rethinking} to reduce the computational burden and memory cost.

\subsubsection{Normal cell}
As shown in the bottom right part of Figure~\ref{fig:structure}, NCs share a similar structure with the RC except for the absence of the PRM, which can provide rich spatial information via aggregating multi-scale information and compensate the spatial information loss caused by downsampling in RC. Given the features containing the multi-scale information, NCs are expected to focus on modeling long- and short-range dependency among the features. Besides, omitting the PRM module in NCs also helps to reduce the computational cost due to the large number of NCs in the stacked models. Therefore, we do not use PRM in NC. Specifically, given $f_{3}$ from the third RC, we first concatenate it with the class token $t_{cls}$, and then add it to the positional encodings to get the input tokens $t$ for the following NCs. We ignore the subscript for clarity since all NCs have an identical architecture but different learnable weights. $t_{cls}$ is randomly initialized at the start of training and fixed during the inference. Similar to the RC, the tokens are fed into the MHSA module, \ie, $t_g = MHSA(t)$. Meanwhile, they are reshaped to 2D feature maps and fed into the PCM, \ie, $t_l = Img2Seq(PCM(Seq2Img(t)))$. Note that the class token is discarded in PCM because it has no spatial connections with other visual tokens.
To further reduce the parameters in NCs, we use group convolutions in PCM. The features from MHSA and PCM are then fused via element-wise sum, \ie, $t_{lg} = t_g + t_l$. 
Finally, $t_{lg}$ are fed into the FFN to get the output features of NC, \ie, $t_{nc} = FFN(t_{lg}) + t_{lg}$.
Similar to ViT~\citep{dosovitskiy2020image}, we apply layer normalization to the class token generated by the last NC and feed it to the classification head to get the final classification result. 

\subsection{Scaling up ViTAE via self-supervised learning}

Except stacking the proposed RCs and NCs to construct the isotropic ViTAE models with 4M, 6M, 13M, and 24M parameters, we also scale up ViTAE to evaluate the benefit of introducing the inductive bias in vision transformers with large model sizes. Specifically, we follow the setting in ViT~\citep{dosovitskiy2020image} to scale up the proposed ViTAE model, \ie, we embed the image into visual tokens and process them using stacked NCs to extract features. The stacking strategy is exactly the same as the strategy adopted in ViT~\citep{dosovitskiy2020image}, where we use 12 NCs with 768 embedding dimensions to construct the ViTAE base model (\ie, ViTAE-B with 89M parameters), 24 NCs with 1,024 embedding dimensions to construct the ViTAE large model (\ie, ViTAE-L with 311M parameters), and 36 normal cells with 1,248 embedding dimension to construct the ViTAE huge model (\ie, ViTAE-H with 644M parameters). The normal cells are stacked sequentially. However, the scaled-up models are easy to overfit if only trained using the ImageNet-1K dataset under a fully supervised training setting. Self-supervised learning~\citep{he2021masked}, on the contrary, can eliminate this issue and facilitate the training of scaled-up models. In this paper, we adopt MAE~\citep{he2021masked} to train the scaled-up ViTAE model due to its simplicity and efficiency. Specifically, we first embed the input images into tokens and then randomly remove 75\% of the tokens. The removed tokens are filled with randomly initialized mask tokens. After that, the remained visual tokens are processed by the ViTAE model for feature extraction. The extracted features and the mask tokens are then concatenated and fed into the decoder network to predict the values of the pixels belonging to the masked regions. The mean squared errors between the prediction and the masked pixels are minimized during the training. 

However, as the encoder only processes the visual tokens, \ie, the remained tokens after removing, the built-in locality property of images has been broken among the visual tokens. To adapt the proposed ViTAE model to the self-supervised task, we simply use convolution with kernel size $1 \times 1$ instead of $3 \times 3$ to formulate the ViTAE model for pretraining. This simple modification helps us to preserve a similar architecture between the network's pretraining and finetuning stage and helps the convolution branch to learn a meaningful initialization, as demonstrated in~\citep{zhang2018fully}. After the pretraining stage, we convert the kernels of the convolutions from $1\times 1$ to $3 \times 3$ by zero-padding to recover the complete ViTAE models, which are further finetuned on the ImageNet-1k training data for 50 epochs. Inspired by~\citep{beit}, we use layer-wise learning rate decay during the finetuning to adapt the pre-trained models for specific vision tasks.  

\subsection{The multi-stage design for ViTAE}

\begin{figure*}[ht]
    \centering
    \includegraphics[width=1.0\linewidth]{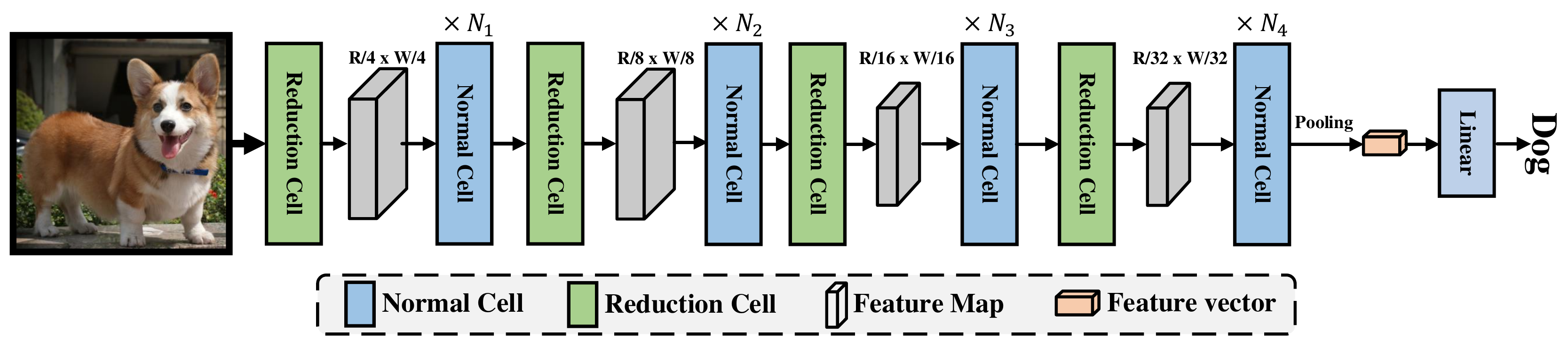}
    \caption{The structure of the proposed ViTAEv2 in the multi-stage manner. The RCs and NCs are re-arranged in a stage-wise manner. At each stage, a number of NCs are sequentially stacked following each RC, which gradually downsamples the features by a certain ratio, \ie, 4$\times$, 2$\times$, 2$\times$, and 2$\times$, respectively.}
    \label{fig:stage-wise structure}
\end{figure*}

Apart from classification, other downstream tasks, including object detection, semantic segmentation, and pose estimation, are also very important that a general backbone should adapt to. These downstream tasks usually need to extract multi-level features from the backbone to deal with those objects at different scales. To this end, we extend the vanilla ViTAE model to the multi-stage design, \ie, ViTAE-v2. A natural choice for the design of ViTAE-v2 can be re-constructing the model by re-organizing RCs and NCs. As shown in Figure~\ref{fig:stage-wise structure}, ViTAE-v2 has four stages where four corresponding RCs are used to gradually downsample the features by $4\times$, $2\times$, $2\times$, and $2\times$, respectively. At each stage, a number of $N_i$ normal cells are sequentially stacked following the $i_{th}$ RC. Note that a series of NCs are used only at the most coarse stage in the isotropic design. The number of normal cells, \ie, $N_i$, controls the model depth and size. By doing so, ViTAE-v2 can extract a feature pyramid from different stages which can be used by the decoders specifically designed for various downstream tasks.

One remaining issue is that the vanilla attention operations in transformers have a quadratic computational complexity, therefore requiring a large memory footprint and computation cost, especially for feature maps with a large resolution. In contrast to the fast resolution reduction in the vanilla ViTAE design, we adopt a slow resolution reduction strategy in the multi-stage design, \eg, the resolution of the feature maps at the first stage is only $1/4$ of the original image size, thereby incurring more computational cost especially when the images in downstream tasks have high resolutions. To mitigate this issue, we further explore another inductive bias, \ie, local window attention introduced in~\citep{liu2021swin}, in the RC and NC modules. Specifically, the window attention split the whole feature map into several non-overlap local windows and conducts the multi-head self-attention within each window, \ie, each \textit{query} token within the same window shares the same \textit{key} and \textit{value} sets. Since the parallel convolution branch in the proposed two cells can encode position information and enable inter-window information exchange, special designs like the relative position encoding and window-shifting mechanism in~\citep{liu2021swin} can be omitted. We empirically find that replacing the full attention with local window attention at early stages can achieve a good trade-off between computational cost and performance. Therefore, we only use local window attention in the RC and NC modules at the first two stages. Consequently, our ViTAEv2 models can deliver superior performance for various vision tasks, including image classification, object detection, semantic segmentation, and pose estimation, while keeping a fast inference speed and reasonable memory footprint.

\subsection{Model details}
\begin{table*}[htbp]
  \centering
  \caption{Model details of ViTAE and ViTAEv2 variants. The two rows in the `RC' and `NC' columns denote the specific configurations of RCs and NCs, \ie, the attention type and number of heads, respectively. '-' denotes there is no RC at the corresponding stage. `arrangement' and `embedding' denote the number of NCs and the token embedding size at each stage. 'P', 'T', 'W' represents performer, vanilla transformer, and local window attention, respectively.}
    \begin{tabular}{c|c|cc|c|c|c|c}
    \hline
    Model & Dilation & RC & NC & Arrangement & Embedding & Params (M) & Flops (G) \\
    \hline
    ViTAE-T & [1,2,3,4] &  \makecell[c]{P, P, P \\ 1, 1, 1} & \makecell[c]{F \\ 4} & 0, 0, 7 & 64, 64, 256 & 4.8   & 1.5 \\
    \hline
    ViTAE-6M & [1,2,3,4] &  \makecell[c]{P, P, P \\ 1, 1, 1} & \makecell[c]{F \\ 4} & 0, 0, 10 & 64, 64, 256 & 6.5   & 2.0 \\
    \hline
    ViTAE-13M & [1,2,3,4] &  \makecell[c]{P, P, P \\ 1, 1, 1} & \makecell[c]{F \\ 4} & 0, 0, 11 & 64, 64, 320 & 13.2  & 3.3 \\
    \hline
    ViTAE-S & [1,2,3,4] &   \makecell[c]{P, P, P \\ 1, 1, 1} & \makecell[c]{F \\ 4} & 0, 0, 14 & 96, 192, 384 & 23.6  & 5.6 \\
    \hline
    ViTAE-B & - &   \makecell[c]{- \\ -} & \makecell[c]{F \\ 12} & 0, 0, 12 & 768 & 89.3  & 36.9 \\
    \hline
    ViTAE-L & - &   \makecell[c]{- \\ -} & \makecell[c]{F \\ 16} & 0, 0, 24 & 1024 & 311.7  & 125.8 \\
    \hline
    ViTAE-H & - &   \makecell[c]{- \\ -} & \makecell[c]{F \\ 16} & 0, 0, 32 & 1280 & 644.3 & 335.7 \\
    \hline
    ViTAEv2-S & [1,2,3,4] $\downarrow$ &  \makecell[c]{W, W, F, F \\ 1, 1, 2, 4} & \makecell[c]{W, W, F, F \\ 1, 2, 4, 8} & 2, 2, 8, 2 & 64, 128, 256, 512 & 19.3 & 5.7 \\
    \hline
    ViTAEv2-48M & [1,2,3,4] $\downarrow$ &  \makecell[c]{W, W, F, F \\ 1, 1, 2, 4} & \makecell[c]{W, W, F, F \\ 1, 2, 4, 8} & 2, 2, 11, 2 & 96, 192, 384, 768 & 48.7 & 13.3 \\
    \hline
    ViTAEv2-B & [1,2,3,4] $\downarrow$ &  \makecell[c]{W, W, F, F \\ 1, 1, 2, 4} & \makecell[c]{W, W, F, F \\ 1, 2, 4, 8} & 2, 2, 12, 2 & 128, 256, 512, 1024 & 89.7 & 24.3 \\
    \hline
    \end{tabular}%
  \label{tab:ViTAEDetails}%
\end{table*}%

In this paper, we propose ViTAE and further extend it to the multi-stage version ViTAEv2 as described above. We devise several ViTAE and ViTAEv2 variants in our experiments to be compared with other models with similar model sizes. The details of them are summarized in Table~\ref{tab:ViTAEDetails}. The `dilation' column determines the dilation rate sets $\mathcal{S}$ in each RC. The two rows in the `RC' and `NC' columns denote the specific configurations of RCs and NCs, respectively, where `P', 'W', 'F' refers to Performer~\citep{choromanski2020rethinking}, local window attention, and the vanilla full attention, respectively, and the number in the second rows denotes the number of heads in the corresponding attention module. The `arrangement' column denotes the number of NC at each stage, while the `embedding' denotes the token embedding size at each stage. Specifically, the default convolution kernel size in the first RC is $7\times 7$ with a stride of $4$ and dilation rates from $\mathcal{S}_1=[1, 2, 3, 4]$. In the following two RCs (or three RCs for ViTAEv2), the convolution kernel size is $3 \times 3$ with a stride of $2$ and dilation rates from $\mathcal{S}_2=[1,2,3]$ and $\mathcal{S}_3=[1,2]$ (and $\mathcal{S}_4=[1,2]$ for ViTAEv2), respectively. Since the number of tokens decreases at later stages, there is no need to use large kernels and dilation rates at later stages. PCM in both RCs and NCs comprises three convolutional layers with a kernel size of $3 \times 3$.

\section{Experiments}
\label{sec:ViTAEexperi}
\subsection{Implementation details}
\label{subsec:impl}

Unless explicitly stated, we train and test the proposed ViTAE and ViVTAEv2 model on the ImageNet-1k~\citep{krizhevsky2012imagenet} dataset, which contains about 1.3 million images from 1k classes. The image size during training is set to $224 \times 224$. We use the AdamW~\citep{loshchilov2018decoupled} optimizer with the cosine learning rate scheduler and use the data augmentation strategy exactly the same as T2T~\citep{yuan2021tokens} for a fair comparison regarding the training strategies and the size of models. We use a batch size of 512 for training ViTAE and 1024 for ViTAEv2. The learning rate is set to be proportion to 512 batch size with a base value 5e-4. The results of our models can be found in Table~\ref{tab:ViTAESuppSota}, where all the models are trained for 300 epochs. The models are built on PyTorch~\citep{paszke2019pytorch} and TIMM~\citep{rw2019timm}.

\begin{table*}[htbp]
  \centering
  \footnotesize
  \caption{Comparison with SOTA methods. {$\uparrow$ 384} denotes finetuning the model using images of 384$\times$384 resolution.}
    \setlength{\tabcolsep}{0.005\linewidth}{\begin{tabular}{c|c|ccc|cc|c|c}
    \hline
    \multirow{2}[1]{*}{{Type}} & \multicolumn{1}{c|}{\multirow{2}[1]{*}{{Model}}} & {Params } & {FLOPs} & {Input} & \multicolumn{2}{c|}{{ImageNet}} & {Real} & Venue \\
          &       & {(M)} & {(G)} & {Size} & {Top-1} & {Top-5} & {Top-1} \\
    \hline
    \multirow{10}[2]{*}{CNN} & ResNet-18~\citep{he2016deep} & 11.7  & 1.8   & 224   & 70.3  & 86.7  & 77.3 & CVPR'16 \\
          & ResNet-50~\citep{he2016deep} & 25.6  & 3.8   & 224   & 76.7  & 93.3  & 82.5 & CVPR'16  \\
          & ResNet-101~\citep{he2016deep} & 44.5  & 7.6  & 224   & 78.3  & 94.1  & 83.7 & CVPR'16  \\
          & ResNet-152~\citep{he2016deep} & 60.2  & 11.3  & 224   & 78.9  & 94.4  & 84.1 & CVPR'16  \\
          & EfficientNet-B0~\citep{tan2019efficientnet} & 5.3   & 0.4   & 224   & 77.1  & 93.3  & 83.5 & ICMR'19 \\
          & EfficientNet-B4~\citep{tan2019efficientnet} & 19.3  & 4.2   & 380   & 82.9  & 96.4  & 88.0 & ICMR'19 \\
          & RegNetY-600M~\citep{radosavovic2020designing} & 6.1  & 0.6   & 224   & 75.5  & -     & - & CVPR'20 \\
          & RegNetY-4GF~\citep{radosavovic2020designing} & 20.6  & 4.0   & 224   & 80.0  & -     & 86.4 & CVPR'20 \\
          & RegNetY-8GF~\citep{radosavovic2020designing} & 39.2  & 8.0   & 224   & 81.7  & -     & 87.4 & CVPR'20 \\
    \hline
    \multirow{50}[10]{*}{Transformer} & DeiT-T~\citep{touvron2020training} & 5.7   & 1.3   & 224   & 72.2  & 91.1  & 80.6 & ICML'21 \\
          & DeiT-T\alambic~\citep{touvron2020training} & 5.7   & 1.3   & 224   & 74.5  & 91.9  & 82.1 & ICML'21  \\
          & PiT-Ti~\citep{heo2021rethinking} & 4.9   & 0.7   & 224   & 73.0  & -     & - & ICCV'21 \\
          & LocalViT-T~\citep{li2021localvit} &  5.9 & 1.3 & 224 & 74.8 & - & - & Arxiv'21 \\
          & T2T-ViT-7~\citep{yuan2021tokens} & 4.3   & 1.2   & 224   & 71.7  & 90.9  & 79.7 & ICCV'21 \\
          & {ViTAE-T} & 4.8   & 1.5   & 224   & 75.3  & 92.7  & 82.9 & NeurIPS'21 \\
\cline{2-9}          & CeiT-T~\citep{yuan2021incorporating} & 6.4   & 1.2   & 224   & 76.4  & 93.4  & 83.6 & ICCV'21 \\
          & ConViT-Ti~\citep{d2021convit} & 6.0   & 1.0   & 224   & 73.1  & -     & -  & ICML'21 \\
          & CrossViT-Ti~\citep{chen2021crossvit} & 6.9   & 1.6   & 224   & 73.4  & -     & - & ICCV'21 \\
          & {ViTAE-6M} & 6.5   & 2.0   & 224   & 77.9  & 94.1  & 84.9 & NeurIPS'21 \\
\cline{2-9}          & PVT-T~\citep{wang2021pyramid} & 13.2  & 1.9   & 224   & 75.1  & -     & - & ICCV'21 \\
          & ConViT-Ti+~\citep{d2021convit} & 10.0  & 2.0   & 224   & 76.7  & -     & - & ICML'21 \\
          & PiT-XS~\citep{heo2021rethinking} & 10.6  & 1.4   & 224   & 78.1  & -     & - & ICCV'21 \\
          & ConT-M~\citep{yan2021contnet} & 19.2  & 3.1   & 224   & 80.2  & -     & - & Arxiv'21 \\
          & 
          XCiT-T24~\citep{el2021xcit} & 12.1 & 2.35 & 224 & 79.4 & - & - & NeurIPS'21 \\
          & PoolFormer-S12~\citep{yu2022metaformer} & 11.9 & 1.82 & 224 & 77.2 & - & - & CVPR'22 \\
          & MPViT-XS~\citep{lee2022mpvit} & 10.5 & 2.97 & 224 & 80.9 &  - & - & CVPR'22 \\
          & VAN-Small~\citep{guo2022visual} & 13.8 & 2.50 & 224 & 81.1 & - & - & Arxiv'22 \\
          & ViTAE-13M & 13.2  & 3.3   & 224   & 81.0  & 95.4  & 86.8 & NeurIPS'21 \\
\cline{2-9}          & DeiT-S~\citep{touvron2020training} & 22.1  & 4.6   & 224   & 79.9  & 95.0  & 85.7 & ICML'21  \\
          & DeiT-S\alambic~\citep{touvron2020training} & 22.1  & 4.6   & 224   & 81.2  & 95.4  & 86.8  & ICML'21 \\
          & Local-ViT~\citep{li2021localvit} & 22.4 & 4.6 & 224 & 80.8 & - & - & Arxiv'21 \\
          & PVT-S~\citep{wang2021pyramid} & 24.5  & 3.8   & 224   & 79.8  & - & - & ICCV'21\\
          & Conformer-Ti~\citep{peng2021conformer} & 23.5  & 5.2   & 224   & 81.3  & - & - & ICLR'21\\
          & Swin-T~\citep{liu2021swin} & 29.0  & 4.5   & 224   & 81.3  & -     & - & ICCV'21 \\
          & CeiT-S~\citep{yuan2021incorporating} & 24.2  & 4.5   & 224   & 82.0  & 95.9  & 87.3 & ICCV'21 \\
          & CvT-13~\citep{wu2021cvt} & 20.0  & 4.5   & 224   & 81.6  & -     & 86.7 & ICLR'21 \\
          & ConViT-S~\citep{d2021convit} & 27.0  & 5.4  & 224   & 81.3  & -     & - & ICML'21 \\
          & CrossViT-S~\citep{chen2021crossvit} & 26.7  & 5.6  & 224   & 81.0  & -     & - & ICCV'21 \\
          & PiT-S~\citep{heo2021rethinking} & 23.5  & 4.8   & 224   & 80.9  & -     & - & ICCV'21 \\
          & TNT-S~\citep{han2021transformer} & 23.8  & 5.2  & 224   & 81.3  & 95.6  & - & NeurIPS'21 \\
          & Twins-PCPVT-S\citep{chu2021twins} & 24.1  & 3.8  & 224   & 81.2  & -  & - & NeurIPS'21 \\
          & Twins-SVT-S~\citep{chu2021twins} & 24.0 & 2.9 & 224 & 81.7 & - & - & NeurIPS'21 \\
          & T2T-ViT-14~\citep{yuan2021tokens} & 21.5  & 5.2   & 224   & 81.5  & 95.7  & 86.8 & ICCV'21 \\
          & XCiT-S12~\citep{el2021xcit} & 26.2 & 18.92 & 224 & 82.0 - & - & - & NeurIPS'21 \\
          & Crossformer-T~\citep{wang2021crossformer} & 27.7 & 2.86 & 224 & 81.5 & - & - & ICLR'22 \\
          & PoolFormer-S36~\citep{yu2022metaformer} &  30.8 & 5.00 & 224 & 81.4 & - & - & CVPR'22 \\
          & DAT-T~\citep{xia2022vision} &  28.3 & 4.58 & 224 & 82.0 & - & - &  CVPR'22 \\
          & ViTAE-S & 23.6  & 5.6   & 224   & 82.0  & 95.9  & 87.0 & NeurIPS'21 \\
          & {ViTAE-S $\uparrow$ 384} & 23.6   & 20.2   & 384   & 83.0  &  96.2  & 87.5 & NeurIPS'21 \\
          & {ViTAEv2-S} &  19.2  &  5.7 &  224 &    82.6    & 96.2  & 87.6 & - \\
          & {ViTAEv2-S $\uparrow$ 384} &  19.2  &  17.8  &  384 &    83.8    & 96.7  & 88.3 & - \\
    \hline
    \end{tabular}}
  \label{tab:ViTAESuppSota}%
\end{table*}%

\begin{table*}[htbp]
  \centering
  \caption{Comparison with SOTA methods (Table~\ref{tab:ViTAESuppSota} continued). {$\uparrow$ 384} denotes finetuning the model using images of 384$\times$384 resolution, while {*} denotes the model pretrained using ImageNet-22k.}
    \setlength{\tabcolsep}{0.005\linewidth}{\begin{tabular}{c|c|ccc|cc|c|c}
    \hline
    \multirow{2}[1]{*}{{Type}} & \multicolumn{1}{c|}{\multirow{2}[1]{*}{{Model}}} & {Params } & {FLOPs} & {Input} & \multicolumn{2}{c|}{{ImageNet}} & {Real} & Venue \\
          &       & \multicolumn{1}{c}{{(M)}} & \multicolumn{1}{c}{{(G)}} & \multicolumn{1}{c|}{{Size}} & \multicolumn{1}{c}{{Top-1}} & \multicolumn{1}{c|}{{Top-5}} & \multicolumn{1}{c|}{{Top-1}} \\
    \hline
    \multirow{10}[1]{*}{Transformer} 
          & ViT-B/16~\citep{dosovitskiy2020image} & 86.5  & 35.1  & 384   & 77.9  & -     & -  & ICLR'21 \\
          & ViT-L/16~\citep{dosovitskiy2020image} & 304.3  & 122.9  & 384   & 76.5  & -     & - & ICLR'21 \\
          & DeiT-B~\citep{touvron2020training} & 86.6  & 17.5  & 224   & 81.8  & 95.6  & 86.7 & ICLR'21 \\
          & PVT-M~\citep{wang2021pyramid} & 44.2  & 13.2  & 224   & 81.2  & -     & - & ICCV'21 \\
          & PVT-L~\citep{wang2021pyramid} & 61.4  & 9.8  & 224   & 81.7  & -     & - & ICCV'21 \\
          & Conformer-S~\citep{peng2021conformer} & 37.7  & 10.6  & 224   & 83.4  & -     & - & ICLR'22 \\
          & Swin-S~\citep{liu2021swin} & 50.0  & 8.7  & 224   & 83.0  & - & - & ICCV'21 \\
          & ConT-B~\citep{yan2021contnet} & 39.6  & 6.4  & 224   & 81.8  & -     & - & Arxiv'21 \\
          & CvT-21~\citep{wu2021cvt} & 32.0  & 7.2  & 224   & 82.5  & -     & 87.2 & ICLR'21 \\
          & ConViT-S+~\citep{d2021convit} & 48.0  & 10.0  & 224   & 82.2  & -     & - & ICML'21 \\
          & ConViT-B~\citep{d2021convit} & 86.0  & 17.0  & 224   & 82.4  & -     & - & ICML'21 \\
          & ConViT-B+~\citep{d2021convit} & 152.0  & 30.0  & 224   & 82.5  & -     & - & ICML'21 \\
          & PiT-B~\citep{heo2021rethinking} & 73.8  & 12.5  & 224   & 82.0  & -     & - & ICCV'21 \\
          & TNT-B~\citep{han2021transformer} & 65.6  & 14.1  & 224   & 82.8  & 96.3  & - & NeurIPS'21 \\
          & T2T-ViT-19~\citep{yuan2021tokens} & 39.2  & 8.9   & 224   & 81.9  & 95.7  & 86.9 & ICCV'21 \\
        & ViL-Base~\citep{zhang2021multi} & 55.7  &  13.4 & 224   & 83.2  & - & - & ICCV'21 \\
        & PoolFormer-M48~\citep{yu2022metaformer} &  73.4 & 11.59 & 224 & 82.5 & - & - & CVPR'22 \\
          & {ViTAEv2-48M} &    48.5   &    13.3   & 224   & 83.8 & 96.6 & 88.4 & - \\
          & {ViTAEv2-48M $\uparrow$ 384} &  48.5   &   41.1    & 384   & 84.7 & 97.0 & 88.8 & -  \\
        \cline{2-9}
          & Swin-B~\citep{liu2021swin} & 88.0  & 15.4 & 224   & 83.3  & - & - & ICCV'21 \\
          & Twins-SVT-L~\citep{chu2021twins} & 99.2  & 14.8 & 224   & 83.7  & - & - & NeurIPS'21 \\
          & PVTv2-B5~\citep{wang2021pvtv2} & 82.0  & 11.8 & 224   & 83.8  & - & - & CVMJ'22 \\
          & Focal-B~\citep{yang2021focal} & 89.8  & 16.0 & 224   & 83.8  & - & - &  NeurIPS'21 \\
          & DAT-B~\citep{xia2022vision} & 88.0  & 15.4 & 224   & 84.0  & - & - & CVPR'22 \\
          & CrossFormer~\citep{wang2021crossformer} & 92.0  & 16.1 & 224   & 84.0  & - & - & ICLR'22 \\
          & Conformer-B~\citep{peng2021conformer} & 83.3  & 23.3 & 224   & 84.1  & - & - & ICCV'21 \\
          & {ViTAEv2-B} & 89.7  &   24.3    & 224   & 84.6  & 96.9  & 88.7 & - \\
          & {ViTAEv2-B $\uparrow$ 384} & 89.7  &     74.4  & 384   & 85.3  & 97.1  & 89.2 & - \\
          & {ViTAEv2-B*} & 89.7  &  24.3  &   224 &   86.1    & 97.9 & 89.9 & -\\
          \hline
    \end{tabular}}%
  \label{tab:ViTAESuppSota1}%
\end{table*}%

\subsection{Comparison with the state-of-the-art}

We compare our ViTAE and ViTAEv2 with both CNN models and vision transformers with similar model sizes in Table~\ref{tab:ViTAESuppSota} and Table~\ref{tab:ViTAESuppSota1}. Both Top-1/5 accuracy and real Top-1 accuracy~\citep{beyer2020we} on the ImageNet validation set are reported. We categorize the methods into CNN models, vision transformers with learned IB, and vision transformers with introduced intrinsic IB. Compared with CNN models, our ViTAE-T achieves a 75.3\% Top-1 accuracy, which is better than ResNet-18 with more parameters. The real Top-1 accuracy of the ViTAE model is 82.9\%, which is comparable to ResNet-50 that has four more times of parameters than ours. Similar phenomena can also be observed when comparing ViTAE-T with MobileNetV1~\citep{howard2017mobilenets} and MobileNetV2~\citep{sandler2018mobilenetv2}, where ViTAE obtains better performance with fewer parameters. ViTAE-S achieves 82.0\% Top-1 accuracy with half of the parameters of ResNet-101 and ResNet-152, showing the superiority of learning both local and long-range features from specific structures with corresponding intrinsic IBs by design. When adopting the multi-stage design, ViTAEv2-S further improves the Top-1 accuracy to 82.6\% significantly. When finetuning the model using images of a larger resolution, \eg, using $384 \times 384$ images as input, ViTAE-S's performance is further improved significantly by 1.2\% absolute Top-1 accuracy. ViTAEv2-48M in Table~\ref{tab:ViTAESuppSota1} also benefits from it, and the performance increases from 83.8\% to 84.7\%, which further shows the potential of vision transformers with intrinsic IBs for large resolution images that are common in downstream dense prediction tasks. When the model size increases to 88M, ViTAEv2-B reaches 84.6\% Top-1 accuracy, significantly outperforming other transformer models including Swin-B~\citep{liu2021swin}, Focal-B~\citep{yang2021focal}, and CrossFormer-B~\citep{wang2021crossformer}. When finetuning using images of larger resolution or pretraining the model with ImageNet-22k, ViTAEv2-B's performance increases to 85.3\% and 86.1\% Top-1 accuracy, respectively, confirming the scalability of using IBs for large models and trained on large-scale datasets.

\subsection{Analysis of the isotropic design of ViTAE}
\subsubsection{Data efficiency and training efficiency}

To validate the effectiveness of introducing intrinsic IBs in improving data efficiency and training efficiency, we compare our ViTAE-T model with the baseline model T2T-ViT-7 at different training settings:
 (a) training them using 20\%, 60\%, and 100\% ImageNet training set for equivalent 100 epochs regarding the full ImageNet training set, \eg, we employ 5 times epochs when using 20\% data for training compared with using 100\% data; 
and (b) training them using the full ImageNet training set for 100, 200, and 300 epochs, respectively. 
The results are shown in Figure~\ref{fig:efficiency}. As can be seen, ViTAE-T consistently outperforms the T2T-ViT-7 baseline by a large margin in terms of both data efficiency and training efficiency. For example, ViTAE-T using only 20\% training data achieves comparable performance with T2T-ViT-7 using all data. When 60\% training data are used, ViTAE-T significantly outperforms T2T-ViT-7 using all data by about an absolute 3\% accuracy. It is also noteworthy that ViTAE-T trained for only 100 epochs has outperformed T2T-ViT-7 trained for 300 epochs. After training ViTAE-T for 300 epochs, its performance is significantly boosted to 75.3\% Top-1 accuracy. With the proposed RCs and NCs, the transformer layers in our ViTAE only need to focus on modeling long-range dependencies, leaving the locality and multi-scale context modeling to its convolution counterparts, \ie, PCM and PRM. Such a ``divide-and-conquer'' strategy facilitates ViTAE's training, making learning more efficient with less training data and fewer training epochs.

\subsubsection{Generalization on downstream classification tasks}
\begin{table*}[htbp]
\footnotesize
  \centering
  \caption{Generalization of ViTAE and SOTA methods on different downstream image classification tasks.}
    \setlength{\tabcolsep}{0.01\linewidth}{\begin{tabular}{c|c|ccccccc}
    \hline
    {Model} & \multicolumn{1}{c|}{{Params (M)}} & \multicolumn{1}{c}{{Cifar10}} & \multicolumn{1}{c}{{Cifar100}} & \multicolumn{1}{c}{{iNat19}} & \multicolumn{1}{c}{{Cars}} & \multicolumn{1}{c}{{Flowers}} & \multicolumn{1}{c}{{Pets}} \\
    \hline
    Grafit ResNet-50~\citep{touvron2020grafit} & 25.6 & - & - & 75.9  & 92.5  & 98.2  & - \\
    EfficientNet-B5~\citep{tan2019efficientnet} & 30 & 98.1  & 91.1  & - & - & 98.5  & - \\
    ViT-B/16~\citep{dosovitskiy2020image} &  86.5  & 98.1  & 87.1  & -     & -     & 89.5  & 93.8 \\
    ViT-L/16~\citep{dosovitskiy2020image} &  304.3 & 97.9  & 86.4  & -     & -     & 89.7  & 93.6 \\
    DeiT-B~\citep{touvron2020training} & 86.6 & 99.1  & 90.8 & 77.7  & 92.1  & 98.4  & - \\
    T2T-ViT-14~\citep{yuan2021tokens} & 21.5 & 98.3 &  88.4  &  -  & -  & -  & - \\
    \hline ViTAE-T & 4.8 & 97.3 & 86.0  & 73.3 &  89.5  & 97.5 &  92.6 \\
    ViTAE-S & 23.6 & 98.8 & 90.8 & 76.0 & 91.4 & 97.8 & 94.2 \\
    \hline
    \end{tabular}}%
  \label{tab:transfer}%
\end{table*}%

We further investigate the generalization of the proposed ViTAE models pre-trained on ImageNet-1k for downstream image classification tasks by finetuning them further on the training sets of several fine-grained classification tasks, including Flowers~\citep{Nilsback08}, Cars~\citep{KrauseStarkDengFei-Fei_3DRR2013}, Pets~\citep{parkhi12a}, and iNaturalist19. We also finetune the proposed ViTAE models pre-trained on ImageNet-1k further on Cifar10~\citep{krizhevsky2009learning} and Cifar100~\citep{krizhevsky2009learning}. The results are shown in Table~\ref{tab:transfer}. It can be seen that ViTAE achieves SOTA performance on most of the datasets using comparable or fewer parameters. These results demonstrate the good generalization ability of our ViTAE models.

\subsubsection{Ablation study of the design of RC and NC}
We use T2T-ViT~\citep{yuan2021tokens} as our baseline model in the following ablation study of our ViTAE. As shown in Table~\ref{tab:ViTAEAblation}, we investigate the hyper-parameter settings of RC and NC in the ViTAE-T model by isolating them separately. All the models are trained for 100 epochs on ImageNet-1k, following the same training setting and data augmentation strategy as described in Section~\ref{subsec:impl}.

We use $\checkmark$ and $\times$ to denote whether or not the corresponding module is enabled during the experiments. If all columns under the RC and NC are marked $\times$ as shown in the first row, the model becomes the standard T2T-ViT-7 model. ``Pre'' denotes the early fusion strategy that fuses output features of PCM and MHSA before FFN, while ``Post'' denotes a late fusion strategy alternatively. The \checkmark in ``BN'' denotes PCM uses BN. ``$\times 3$'' in the first column denotes that the dilation rate set is the same in the three RCs. ``$[1,2,3,4] \downarrow$'' denotes using smaller dilation rates in deeper RCs, \ie, $\mathcal{S}_1=[1,2,3,4]$, $\mathcal{S}_2=[1,2,3]$, $\mathcal{S}_3=[1,2]$.

\begin{table}
  \centering
  \caption{Ablation Study of RC and NC in ViTAE-T. ``Pre'' denotes the early fusion strategy that fuses output features of PCM and MHSA before FFN while ``Post'' denotes a late fusion strategy alternatively. The \checkmark in ``BN'' denotes PCM uses BN. ``$\times 3$'' in the first column denotes that the dilation rate set is the same in the three RCs. ``$[1,2,3,4] \downarrow$'' denotes using smaller dilation rates in deeper RCs, \ie, $\mathcal{S}_1=[1,2,3,4]$, $\mathcal{S}_2=[1,2,3]$, $\mathcal{S}_3=[1,2]$.}
  \small
    \setlength{\tabcolsep}{0.02\linewidth}{\begin{tabular}{cc|ccc|r}
    \hline
    \multicolumn{2}{c|}{{Reduction Cell}} & \multicolumn{3}{c|}{{Normal Cell}} & \multicolumn{1}{c}{\multirow{2}[4]{*}{{Top-1}}} \\
\cmidrule{1-5}  {Dilation ($\mathcal{S}_1\sim\mathcal{S}_3$)} & {PCM} & {Pre}   & {Post}  & {BN}    &  \\
    \hline
     ×     & ×     & ×     & ×  & ×    & 68.7  \\
    \hline
     ×     & ×     & \checkmark     & ×     & ×     &  69.1 \\
     ×     & ×     & ×     & \checkmark     & ×     &  69.0 \\ 
     ×     & ×     & ×     & \checkmark     & \checkmark & 68.8 \\
     ×     & ×     & \checkmark     & ×     & \checkmark & 69.9 \\
    \hline
     $[1,2] \times 3$  & ×  & ×   & ×       & ×     &  69.5 \\
     $[1,2,3] \times 3$  & ×  & ×   & ×       & ×     &  69.9 \\
     $[1,2,3,4] \times 3$  & ×  & ×   & ×      & ×    & 69.2 \\
     $[1,2,3,4,5] \times 3$  & ×  & ×   & ×      & ×     & 68.9 \\
     $[1,2,3,4] \downarrow$ & ×  & ×      & ×     & ×     & 69.8 \\
     $[1,2,3,4] \downarrow$ & \checkmark  & ×      & ×     & ×     & 71.7 \\
    \hline
     $[1,2,3,4] \downarrow$ & \checkmark     & \checkmark     & ×     & \checkmark     & 72.6 \\
    \hline
    \end{tabular}}%
  \label{tab:ViTAEAblation}
\end{table}%
As can be seen, using an early fusion strategy and BN in NC achieves the best 69.9\% Top-1 accuracy among other settings. It is noteworthy that all the variants of NC outperform the vanilla T2T-ViT, implying the effectiveness of PCM, which introduces the intrinsic locality IB in transformers. It can also be observed that BN plays an important role in improving the model's performance as it can help to alleviate the scale deviation between convolution's and attention's features. For RC, we first investigate the influence of using different dilation rates in the PRM, as shown in the first column. As can be seen, using larger dilation rates (\eg, 4 or 5) does not deliver better performance. We suspect that larger dilation rates may lead to plain features in the deeper RCs due to the smaller resolution of feature maps. To validate the hypothesis, we use smaller dilation rates in deeper RCs as denoted by $[1, 2, 3, 4]\downarrow$. As can be seen, it achieves comparable performance as $[1, 2, 3]\times$. However, compared with $[1, 2, 3, 4]\downarrow$, $[1, 2, 3]\times$ increases the amount of parameters from 4.35M to 4.6M. Therefore, we select $[1, 2, 3, 4]\downarrow$ as the default setting. In addition, using PCM in the RC introduces the intrinsic locality IB and the performance increases to 71.7\% Top-1 accuracy. Finally, the combination of RCs and NCs achieves the best accuracy at 72.6\%, demonstrating their complementarity.

\subsubsection{Visual inspection of ViTAE}

\begin{figure*}
    \centering
    \includegraphics[width=\linewidth]{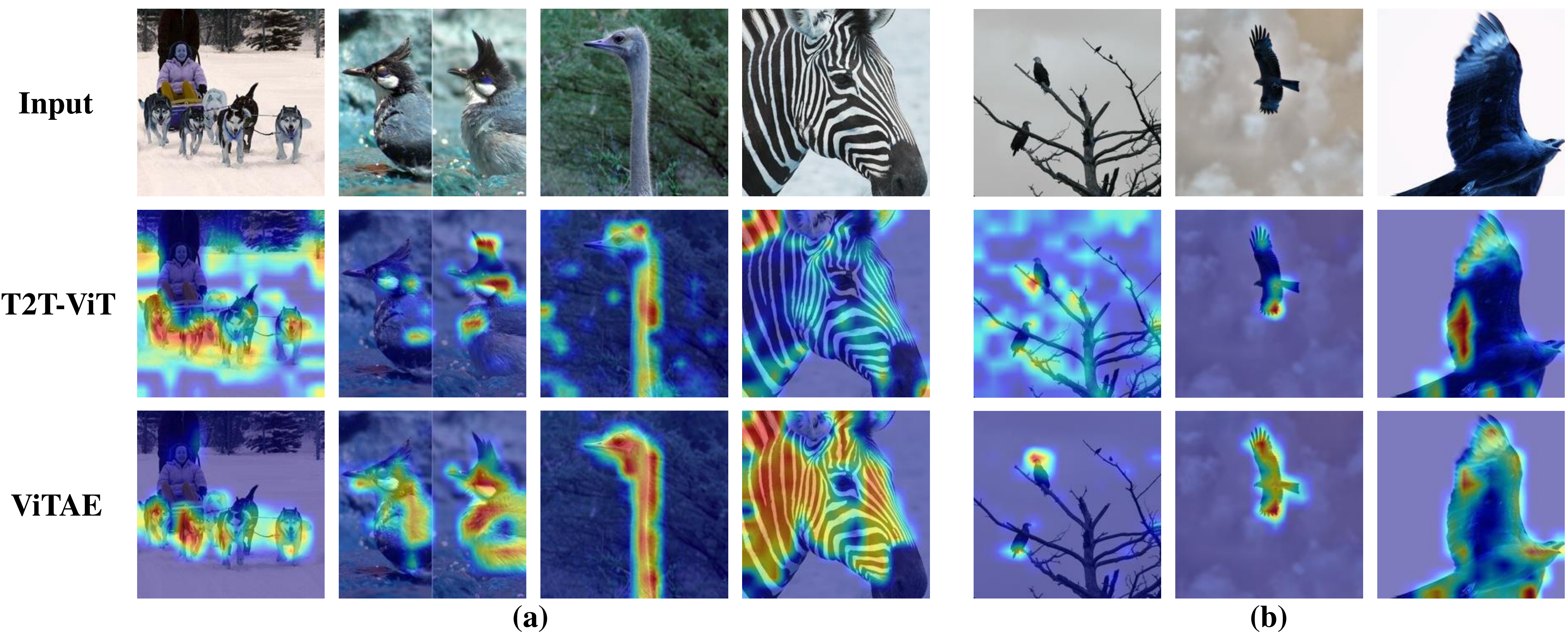}
    \caption{Visual inspection of T2T-ViT-7 and ViTAE-T using Grad-CAM \citep{selvaraju2017grad}. (a) Images containing multiple or single objects and the heatmaps obtained by T2T-ViT-7 and ViTAE-T. (b) Images containing the same class of objects at different scales and the heatmaps obtained by T2T-ViT-7 and ViTAE-T. Best viewed in color.}
    \label{fig:visualInspect}
\end{figure*}

\begin{figure}
    \includegraphics[width=\linewidth]{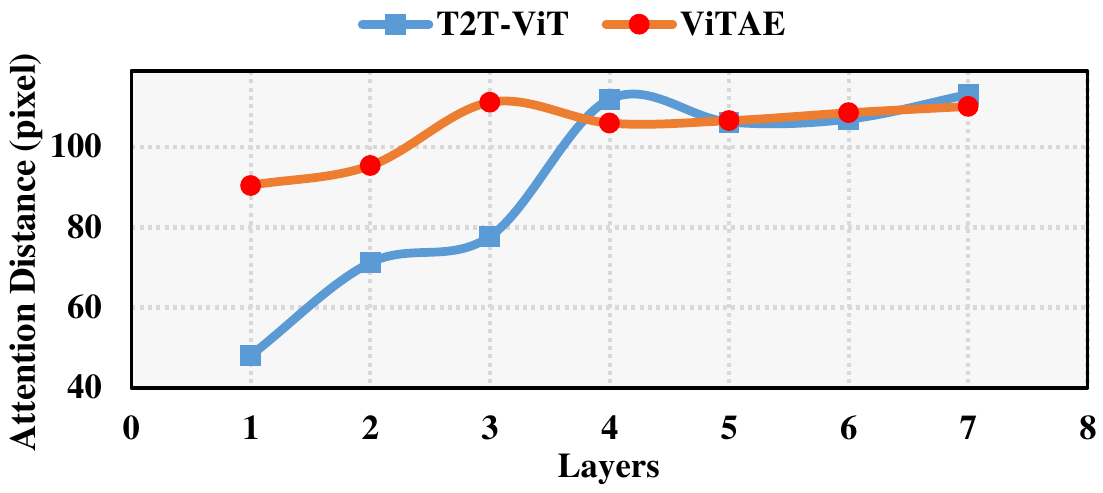}
    \caption{The average per-layer attention distance of T2T-ViT-7 and our ViTAE-T on the ImageNet validation set.}
    \label{fig:attentionDistance}
\end{figure}

To further analyze the property of our ViTAE, we apply Grad-CAM~\citep{selvaraju2017grad} on the MHSA's output in the last NC to qualitatively inspect ViTAE. The visualization results are provided in Figure~\ref{fig:visualInspect}. Compared with the baseline T2T-ViT, our ViTAE covers the single or multiple targets in the images more precisely and attends less to the background. Moreover, ViTAE can better handle the scale variance issue as shown in Figure~\ref{fig:visualInspect}(b). That is, it covers birds accurately whether they are small, medium, or large in size. Such observations demonstrate that introducing the intrinsic IBs of locality and scale-invariance from convolutions to transformers helps ViTAE learn more discriminate features than the pure transformers.

Besides, we calculate the average attention distance of each layer in ViTAE-T and the baseline T2T-ViT-7 on the ImageNet validation set, respectively. The results are shown in Figure~\ref{fig:attentionDistance}. It can be observed that with the usage of PCM, which focuses on modeling locality, the transformer layers in the proposed NCs can better focus on modeling long-range dependencies, especially in shallow layers. In the deep layers, the average attention distances of ViTAE-T and T2T-ViT-7 are almost the same, where modeling long-range dependencies is much more important. It implies that the PCM does not affect the transformer's behavior in deep layers. These results confirm the effectiveness of the adopted ``divide-and-conquer'' idea in ViTAE, \ie, introducing the intrinsic locality IB from convolutions into vision transformers makes it possible that transformer layers only need to be responsible to long-range dependencies since convolutions can well model locality in PCM. 

\subsection{Analysis of the scaled up ViTAE models}
\subsubsection{Image classification performance}

\begin{table}[htbp]
  \centering
  \footnotesize
  \caption{The performance of scaled up ViTAE models on the ImageNet1K dataset. $^*$ denotes the results that we re-implement on GPU with PyTorch framework. $^\dag$ indicates that ImageNet22K are used to further finetune the models with 224$\times$224 resolution for 90 epochs. `Sup' is the short for supervised learning. }
    \setlength{\tabcolsep}{0.01\linewidth}{\begin{tabular}{l|ccccc}
    \hline
           & {\#Params} & \makecell[c]{Test \\ size} & {Method} & \makecell[c]{ImageNet \\ Top-1} & \makecell[c]{Real \\ Top-1} \\
    \hline
    T2TViT-24 & 65 M & 224 & Sup & 82.3 & 87.2 \\
    ViT-B$^*$ & 88 M  & 224   & MAE   & 83.4 & 89.1 \\
    ViTAE-B & {89 M}  & {224}   & {MAE}   & {83.8} & 89.4 \\
    ViTAE-B$^\dag$ & {89 M}  & {224}   & {MAE}   & {84.8} & 89.9 \\
    \hline
    Swin-L$^\dag$ & 197 M & 384 & Sup & 87.3 & 90.0 \\
    SwinV2-L$^\dag$ & 197 M & 384 & Sup & 87.7 & - \\
    CoAtNet-4$^\dag$ & 275 M & 384 & Sup & 87.9 & - \\
    CvT-W24$^\dag$ & 277 M & 384 & Sup & 87.7 & - \\
    ViT-L$^*$ & 304 M & 224   & MAE   & 85.5 & 90.1 \\
    ViT-L & 304 M & 224   & MaskFeat & 85.7 & - \\
    {ViTAE-L} & {311 M} & {224}   & {MAE}   & {86.0} & 90.3 \\
    {ViTAE-L$^\dag$} & {311 M} & {224}   & {MAE}   & {87.5} & 90.8 \\
    {ViTAE-L$^\dag$} & {311 M} & {384}   & {MAE}   & {88.3} & 91.1 \\
    \hline
    SwinV2-H & 658 M & 224   & SimMIM & 85.7 & - \\
    SwinV2-H & 658 M & 512   & SimMIM & 87.1 & - \\
    {ViTAE-H} & {644 M} & {224}   & {MAE}    & {86.9} & 90.6 \\
    {ViTAE-H} & {644 M} & {512}   & {MAE}    & {87.8} & 91.2 \\
    {ViTAE-H$^\dag$} & {644 M} & {224}   & {MAE}    & {88.0} & 90.7 \\
    {ViTAE-H$^\dag$} & {644 M} & {448}   & {MAE}    & {88.5} & 90.8 \\
    \hline
    \end{tabular}}%
  \label{tab:ViTAEScaled}%
\end{table}%

We evaluate the performance of the scaled-up models on the ImageNet dataset. The scaled-up models are pre-trained for 1600 epochs using MAE~\citep{he2021masked}, taking images from the ImageNet-1K training set. Then, the models are finetuned for 100 (the base model) or 50 (the large and huge model) epochs using the labeled data from the ImageNet-1K training set. It should be noted that the original MAE is trained on the TPU machines with Tensorflow, while our implementation adopts PyTorch as the framework and uses NVIDIA GPU for the training. This implementation difference may cause a slight performance difference in the models' classification accuracy. We compare our methods with T2TViT~\citep{yuan2021tokens}, CvT~\citep{yuan2021incorporating}, Swin~\citep{liu2021swin}, SwinV2~\citep{swinv2}, and ViT~\citep{dosovitskiy2020image} with either supervised learning or self-supervised learning like MAE~\citep{he2021masked}, MaskFeat~\citep{wei2021masked}, and SimMIM~\citep{xie2021simmim} The results are summarized in Table~\ref{tab:ViTAEScaled}. It demonstrates that the proposed ViTAE-B model with the introduced inductive bias outperforms the baseline ViT-B model by 0.4 Top-1 classification accuracy. For the ViTAE-L model with 300M parameters, the inductive bias still brings about 0.3 performance gains. After using ImageNet-22K labeled data for finetuning, the classification accuracy on the ImageNet-1K validation set further increases by about 1\%. These results show that the benefit of introducing inductive bias in vision transformers is scalable to large models and datasets. Notably, our ViTAE-H, trained with only the ImageNet-1K dataset, obtains a classification accuracy of 91.2 on the ImageNet Real dataset~\citep{beyer2020we}, which is the highest accuracy we are aware of. It outperforms other methods trained with additional private data, such as EfficientNet~\citep{Pham_2021_CVPR} and ViT-G~\citep{zhai2021scaling}, where the former obtains 91.1 accuracy using the JFT300M dataset and the latter obtains 90.8 accuracy using the JFT3B dataset.

\subsubsection{Few-shot learning performance}

\begin{figure}
    \centering
    \includegraphics[width=\linewidth]{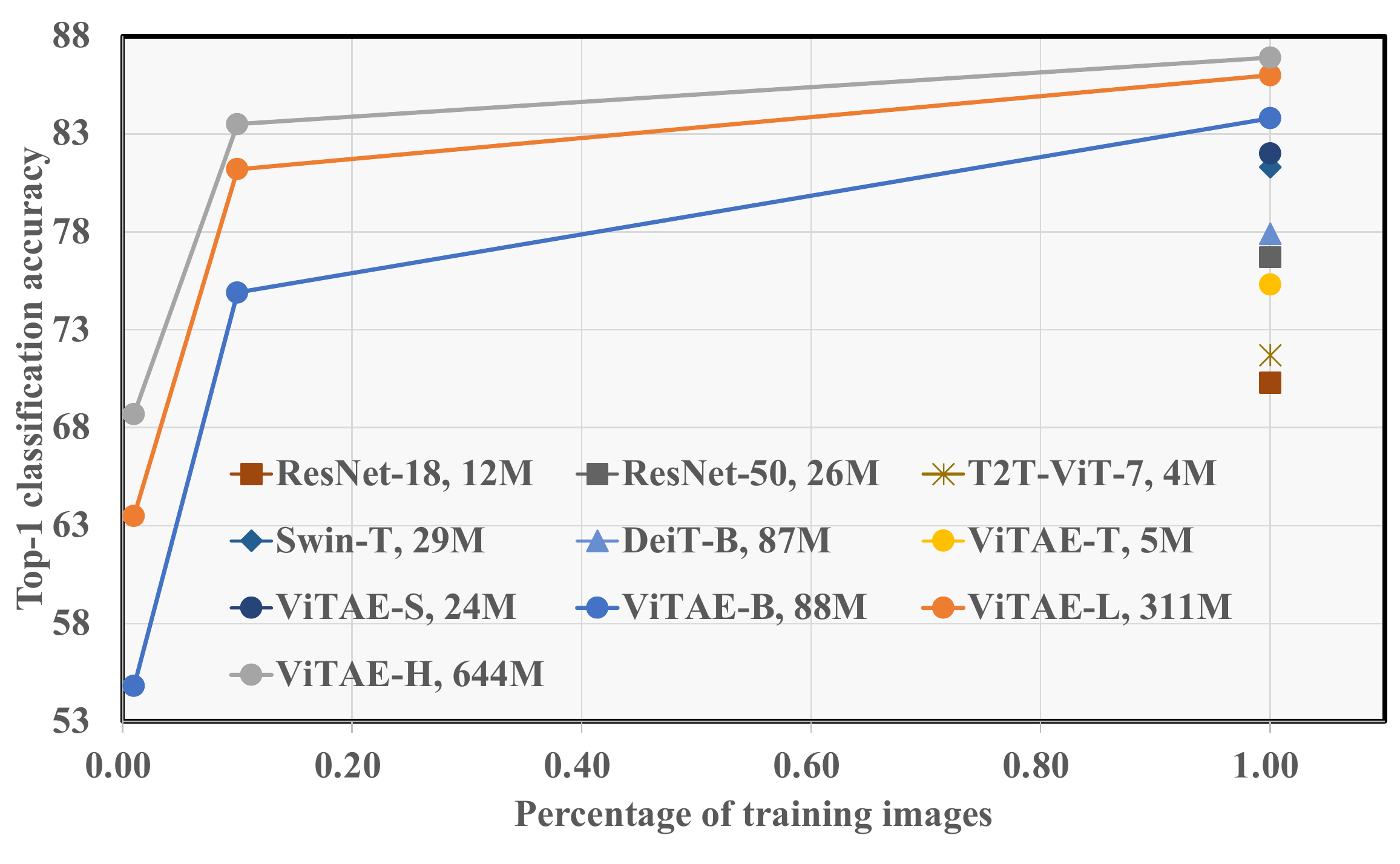}
    \caption{The few-shot image classification results of scaled up ViTAE models. With only 10\% data for finetuning, the scaled up ViTAE models obtain comparable or even better performance compared with the small CNNs and transformer models with 100\% data for training.}
    \label{fig:ViTAEFewShot}
\end{figure}

We further evaluate the data efficiency of scaled-up models by using different percentages of data to finetune the pre-trained models. We use 1\%, 10\%, and 100\% data from the ImageNet-1k training set to finetune the self-supervised pre-trained ViTAE models with different amounts of parameters. We ensure that each model sees the same amount of the training images under different data settings, \ie, we train the ViTAE-B model for 10,000 epochs using 1\% training data, for 1,000 epochs using 10\% training data, and for 100 epochs using 1\% training data. Similarly, the ViTAE-L and ViTAE-H models are trained for 5,000 epochs using 1\% training data and 500 epochs using 10\% training data. The smaller models, \ie, with less than 20M parameters, are trained from scratch using 100\% ImageNet-1k training data for 300 epochs. As shown in Figure~\ref{fig:ViTAEFewShot}, the models with more parameters are more data-efficient than those with fewer parameters. For example, the ViTAE-H model with 644M parameters trained with 10\% data outperforms the small model with 13.2 parameters trained with 100\% data, \ie, 82.4 v.s. 81.0.
Such observations mirror the findings in previous studies, including both image classification and language modeling.

\subsubsection{Ablation study of the convolutional kernel size in the scaled up ViTAE models}

\begin{table}[htbp]
  \centering
   \small
  \caption{The influence of the convolutional kernel size in the scaled up ViTAE models during pretraining.}
    \begin{tabular}{l|cc|c}
    \hline
          & Kernel Size & \#Params & Accuracy \\
    \hline
    ViT-L & 0     & 304 M  & 84.1 \\
    ViTAE-L & 0     & 311 M  & 84.1 \\
    ViTAE-L & 3     & 311 M   & 84.4 \\
    ViTAE-L & 1     & 311 M  & 84.6 \\
    \hline
    \end{tabular}%
  \label{tab:ViTAEAblationKernel}%
\end{table}%

We conduct experiments to investigate the influence of the convolutional kernel size in the scaled-up ViTAE models during pretraining. The results are presented in Table~\ref{tab:ViTAEAblationKernel}. The kernel size 0 represents that we do not use the convolution branch in ViTAE-L during pretraining, which degenerates to the original ViT-L model. Then, we add the convolution branches during finetuning and initialize the convolutional kernel weight as 0. If we use 1 $\times$ 1 kernels in ViTAE-L during pretraining, we pad them to 3 $\times$ 3 with zero padding during finetuning. We pretrain the models for 400 epochs and further finetune them for 50 epochs on the ImageNet-1K training set. As can be seen, using no convolution branch during pretraining leads to no improvement over the baseline ViT-L since those convolutional kernel weights in ViTAE-L during finetuning are zero-initialized. Directly pretraining and finetuning ViTAE-L with 3 $\times$ 3 convolutional kernels in the convolution branches leads to slightly better performance over the baseline ViT-L, but it is inferior to the proposed setting, \ie, using 1 $\times$ 1 convolutional kernels in the convolution branches during pretraining ViTAE-L while zero-padding them to 3 $\times$ 3 during finetuning. We argue the reason is that most tokens (75\%) during pretraining are randomly removed, and the remaining ones have lost spatial information. Therefore, using 3 $\times$ 3 kernels may lead to overfitting while 1 $\times$ 1 convolutions pay little attention to spatial structures and could learn better feature representation, which is in line with the observations in \citep{zhang2018fully}.

\subsection{Analysis of the multi-stage design ViTAEv2}
\subsubsection{Ablation study of multi-stage design}
\label{Stage-wise ViTAE design}

\begin{table*}[htbp]
  \centering
  \caption{Ablation study of the stage-wise ViTAE design. `P', `W', and `F' represent using the Performer attention, window attention, and vanilla attention for each stage respectively. bs is the short name for batch size. We report the memory footprint and images throughout during training for comparison.}
    \begin{tabular}{c|c|c|c|c|c|c}
    \hline
    \multicolumn{2}{c|}{{Block type}} & P, F, F, F & W, F, F, F & W, W, F, F & W, W, W, F & W, W, W, W \\
    \hline
    \multicolumn{2}{c|}{{Params (M)}} & 19.4  & 19.4  & 19.4  & 19.4  & 19.4 \\
    \hline
    \multicolumn{2}{c|}{{Top-1 Accuracy (224 img size)}} & 82.2  & 82.7  & 82.6  & 82.2  & 82.2 \\
    \hline
    {img size 224, } & Memory (GB) & 32.2  & 31.3  & 28.1  & 27.1  & 27.1 \\
    {bs 128 per GPU} & Training speed (img/s) & 1328.8  & 1300.8  & 1377.6  & 1333.6  & 1335.2 \\
    \hline
    {img size 448, } & Memory (GB) &  40.5 & 40.5  & 32.3  & 26.8  & 26.7 \\
    {bs 32 per GPU} & Training speed (img/s) & 297.6 & 295.2 & 338.4 & 342.4   & 344.8 \\
    \hline
    {img size 672, } & Memory (GB) & 38.7  & 38.8  & 23.7  & 16.3  & 16.0 \\
    {bs 8 per GPU} & Training speed (img/s) & 97.4    & 97.6    & 116.8   & 125.2   & 127.3 \\
    \hline
    {img size 896, } & Memory (GB) & 34.4  & 34.3  & 15.0  & 9.5   & 9.2 \\
    {bs 2 per GPU} & Training speed (img/s) & 37.0    & 36.5    & 44.5    & 45.0    & 45.7 \\
    \hline
    \end{tabular}%
  \label{tab:StageWiseArchDesign}%
\end{table*}%

\begin{table}[htbp]
  \centering
  \caption{Ablation study of the window shifting mechanism and relative position encoding in the local window attention in the ViTAEv2-S model.
  }
    \begin{tabular}{c|cccc}
    \hline
    shift & Y     & Y     & N     & N \\
    relative position enc. & Y     & N     & Y     & N \\
    \hline
    Top-1 Accuracy & 82.7  & 82.4  & 82.5  & 82.6 \\
    Memory (GB) & 28.8  & 28.7  & 28.8  & 28.7 \\
    Training time (s/epoch) & 486  & 477  & 481  & 472 \\
    \hline
    \end{tabular}%
  \label{tab:ablation_RPE_Shift}%
\end{table}%

In this paper, we extend ViTAE to a multi-stage design and propose ViTAEv2 accordingly. To achieve a good trade-off between classification performance and computational cost, we study the design choice of the attention type at each stage. The results are summarized in Table~\ref{tab:StageWiseArchDesign}, where `P', `W', `F' refer to the Performer attention, local window attention, and vanilla attention, respectively. They only differ in the implementation of attention calculation while having the same number of parameters. We list the Top-1 classification accuracy of different model variants trained from scratch using 224$\times$224 images from the ImageNet-1k training set. We gradually increase the image resolution to compare the memory footprint and training speed of different models considering that the backbone models should well adapt to downstream vision tasks where large resolution images are common. Specifically, we set the batch size to 128 for all models for the 224$\times$224 resolution and reduce it for larger resolutions to fit the A100 GPU memory. 

We start from a baseline multi-stage design where the performer attention is used at the first stage while the vanilla full attention is used at the following three stages, denoting as `P,F,F,F'. Then, we gradually introduce inductive bias into the model by replacing the performer and full attention with local window attention. As can be seen, all the models with introduced inductive bias outperform or are at least comparable to the baseline in terms of both classification performance and training cost. Specifically, using local window attention at the first two stages (\ie, `W,W,F,F') leads to the best trade-off between classification performance and computational cost for different image resolutions. Compared with the model with the best performance (\ie, `W,F,F,F'), its classification accuracy only drops by 0.1 while the memory footprint is significantly reduced by 56.4\% at the setting 896$\times$896 image resolution. Moreover, it outperforms the other two designs (\ie, `W,W,W,F' and `W,W,W,W') by an absolute 0.4\% Top-1 accuracy while having about the same training speed. Therefore, we choose the design of `W, W, F, F' and devise the ViTAEv2 models at different model sizes accordingly.

One following interesting question is whether we still need the window-shifting mechanism to enable the inter-window information exchange and the relative position encoding (RPE) in the original implementation of local window attention proposed in \citep{liu2021swin} since the convolutional layers in PRM and PCM can enable inter-window information exchange and encode position information. We carry out an ablation study by isolating them one by one in our ViTAEv2 model to answer this question. We choose ViTAEv2-S as the base model, and all model variants are trained using 224$\times$224 images. The results are summarized in Table~\ref{tab:ablation_RPE_Shift}. As can be seen, the above two components only contribute marginally in our ViTAE model, \ie, about 0.1\% accuracy. Therefore, we do not include them in our default design to make the model simple and easy to implement. 

\begin{table}[htbp]
  \centering
  \caption{Inference speed comparison of ViTAEv2 with ViT~\citep{touvron2020training}, T2T-ViT~\citep{yuan2021tokens} and Swin Transformer~\citep{liu2021swin}.}
    \setlength{\tabcolsep}{0.03\linewidth}{\begin{tabular}{c|ccc|c}
    \hline
          & \makecell[c]{bs 128, \\ size 224 \\ (img/s)} & \makecell[c]{bs 64, \\ size 448 \\ (img/s)} & \makecell[c]{bs 16, \\ size 896 \\ (img/s)} & Acc. \\
    \hline
    ViT-Small & 1459  & 318   & 48    & 79.9 \\
    ViT-Base~ & 803 & 167 & 25 & 81.8 \\
    T2T-ViT-14 & 996   & 220   & 33    & 81.2 \\
    T2T-ViT-24 & 575 & 118 & 17 & 82.3 \\
    Swin-T & 815   & 246   & 60    & 81.3 \\
    {ViTAEv2-S} & {722}   & {205}   & {46}    & {82.6} \\
    \hline
    \end{tabular}}%
  \label{tab:inference speed}%
\end{table}%

\begin{table*}[htbp]
  \centering
  \footnotesize
  \caption{Object detection results on the MS COCO~\citep{lin2014microsoft} validation set regarding different backbones. `Sch' is short for training schedules.
  }
    \setlength{\tabcolsep}{0.0002\linewidth}{\begin{tabular}{c|c|c|c|ccc|ccc|c}
    \hline
    Decoder & Sch & Backbone & Venue & $AP^{b}$ & $AP^{b}_{50}$ & $AP^{b}_{75}$ & $AP^{m}$ & $AP^{m}_{50}$ & $mAP^{m}_{75}$ & Params (M) \\
    \hline
    \multirow{17}[4]{*}{Mask RCNN} & \multirow{11}[2]{*}{1x} & ResNet-50~\citep{he2016deep} & CVPR'16 & 38.0    & 58.6  & 41.4  & 34.4  & 55.1  & 36.7  & 44 \\
          &       & PVT-S~\citep{wang2021pyramid} & ICCV'21 & 40.4  & 62.9  & 43.8  & 37.8  & 60.1  & 40.3  & 44 \\
          &       & Swin-T~\citep{liu2021swin} & ICCV'21 & 43.7  & 66.6  & 47.7  & 39.8  & 63.3  & 42.7  & 47 \\
          &       & Focal-T~\citep{yang2021focal} & NeurIPS'21 & 44.8  & 67.7  & 49.2  & 41.0    & 64.7  & 44.2  & 49 \\
      &       & PVTv2-B2~\citep{wang2021pvtv2} & CVMJ'22 & 45.3  & 67.1  & 49.6  & 41.2  & 64.2  & 44.4  & 45 \\
          &       & RegionViT-B~\citep{chen2021regionvit} & ICLR'21  & 43.5  & 66.7  & 47.4  & 40.1  & 63.4  & 43.0    & 92 \\
          &       & Conformer-S/32~\citep{peng2021conformer} & ICCV'21 & 43.6  & -     & -     & 39.7  & -     & -     & 58 \\
          &       & DAT-T~\citep{xia2022vision} & CVPR'22 & 44.4  & 67.6  & 48.5  & 40.4  & 64.2  & 43.1  & 48 \\
          &       & CrossFormer-S~\citep{wang2021crossformer} & ICLR'21  & 45.4  & 68.0    & 49.7  & 41.4  & 64.8  & 44.6  & 50 \\
          &       & {ViTAEv2-S} & - & {46.3} & {68.8} & {51.0} & {41.8} & {65.6} & {44.9} & {37} \\
\cline{2-11} \citep{he2017mask} & \multirow{6}[2]{*}{3x} & ResNet-50~\citep{he2016deep} & CVPR'16 & 41.0    & 61.7  & 44.9  & 37.1  & 58.4  & 40.1  & 44 \\
          &       & PVT-S~\citep{wang2021pyramid} & ICCV'21 & 43.0    & 65.3  & 46.9  & 39.9  & 62.5  & 42.8  & 44 \\
          &       & Swin-T~\citep{liu2021swin} & ICCV'21 & 46.0    & 68.2  & 50.2  & 41.6  & 65.1  & 44.8  & 48 \\
          &       & MViT-T~\citep{fan2021multiscale} & ICCV'21 & 45.9  & 68.7  & 50.5  & 42.1  & 66.0    & 45.4  & 46 \\
          &       & DAT-T~\citep{xia2022vision} & CVPR'22 & 47.1  & 69.2  & 51.6  & 42.4  & 66.1  & 45.5  & 48 \\
          &       & {ViTAEv2-S} & - & {47.8} & {69.4} & {52.2} & {42.6} & {66.6} & {45.8} & {37} \\
    \hline
          & \multirow{5}[2]{*}{1x} & ResNet-50~\citep{he2016deep} & CVPR'16 & 41.2  & 59.4  & 45.0    & 35.9  & 56.6  & 38.4  & 82 \\
          &       & Swin-T~\citep{liu2021swin} & ICCV'21 & 48.1  & 67.1  & 52.2  & 41.7  & 64.4  & 45.0    & 86 \\
          &       & DAT-T~\citep{xia2022vision} & CVPR'22 & 49.1  & 68.2  & 52.9  & 42.5  & 65.4  & 45.8  & 86 \\
    Cascade  &       & {ViTAEv2-S} & - & {50.6} & {69.9} & {54.9} & {43.6} & {66.9} & {47.2} & {75} \\
\cline{2-11}    Mask RCNN & \multirow{5}[2]{*}{3x} & ResNet-50~\citep{he2016deep} & CVPR'16 & 46.3  & 64.3  & 50.5  & 40.1  & 61.7  & 43.4  & 82 \\
    \citep{cai2019cascade}  &       & Swin-T~\citep{liu2021swin} & ICCV'21 & 50.4  & 69.2  & 54.7  & 43.7  & 66.6  & 47.3  & 86 \\
          &       & PVTv2-B2~\citep{wang2021pvtv2} & CVMJ'22 & 51.1  & 69.8  & 55.3  & - & - & - & 83 \\
          &       & DAT-T~\citep{xia2022vision} & CVPR'22 & 51.3  & 70.1  & 55.8  & 44.5  & 67.5  & 48.1  & 86 \\
          &       & {ViTAEv2-S} & - & {51.4} & {70.4} & {55.6} & {44.5} & {67.8} & {48.2} & {75} \\
    \hline
    \end{tabular}}%
  \label{tab:detection results}%
\end{table*}%

We also compare ViTAEv2 and some representative transformer models in terms of inference speed in Table~\ref{tab:inference speed}. All the experiments are conducted on the same A100 GPU, and TensorRT is adopted to accelerate all models. As can be seen, our ViTAEv2-S model outperforms ViT-Small by 2.7\% Top-1 accuracy while keeping a fast inference speed, especially for large size images, \eg, 896$\times$896. Compared with the state-of-the-art Swin transformer, the inference speed of ViTAEv2-S is slightly slower, \ie, about 10\%$\sim$20\%, but its classification performance is significantly improved by an absolute 1.3\% Top-1 accuracy. ViTAEv2-S also outperforms T2T-ViT-24 in terms of both performance and inference speed.

We further evaluate the proposed ViTAEv2 models on representative downstream vision tasks, including object detection, semantic segmentation, and pose estimation. The results are detailed below.

\subsubsection{The performance on object detection and instance segmentation}

\textbf{Settings} To evaluate ViTAEv2's performance on object detection and instance segmentation tasks, we adopt Mask RCNN~\citep{he2017mask} and Cascade RCNN~\citep{cai2018cascade} as the detection framework and finetune the models on COCO 2017 dataset, which contains 118K training images, 5K validation images, and 20K test-dev images. We adopt exactly the same training setting used in Swin~\citep{liu2021swin}, \ie, multi-scale training, AdamW optimizer~\citep{loshchilov2017decoupled} and the mmdetection code base. The models are trained for 12 (the 1x setting) and 36 epochs (the 3x setting), respectively. We compare the performance of ViTAEv2-S and other backbones, including the classic CNNs, \ie, ResNet~\citep{he2016deep}, and current transformer models.

\noindent \textbf{Results} The results are summarized in Table~\ref{tab:detection results} and ViTAEv2-S achieves the best performance with the least number of parameters. Thanks to the introduced inductive bias like locality and scale invariance, the proposed ViTAEv2 model obtains 2.6 $AP^{b}$ and 2.0 $AP^{m}$ performance gains over Swin when using Mask RCNN as the decoder for the 1$\times$ setting. It also significantly outperforms other backbones like Conformer and CrossFormer, owning to our model's efficient divide-and-conquer structure design. When we extend the training schedule to the 3$\times$ setting (36 epochs in total), ViTAEv2 reaches 50.6 $AP^b$ and 42.6 $AP^m$, significantly better than the other models. It is noteworthy that ViTAEv2 trained for 12 epochs has outperformed Swin-T trained for 36 epochs, validating the data efficiency of our model by introducing the inductive bias. The superiority of ViTAEv2 retains when using Cascade RCNN as the decoder, obtaining 50.6 $AP^b$ and 43.6 $AP^m$ when training 12 epochs and 51.4 $AP^b$ and 44.5 $AP^m$ when training 36 epochs. It can be concluded that introducing inductive bias into transformers helps our model better utilize the data and deliver the best performance for both object detection and instance segmentation.

\begin{table}[htbp]
  \centering
  \caption{Semantic segmentation results on the ADE20k~\citep{zhou2017scene} validation set regarding different backbones including ResNet-50~\citep{he2016deep}, Swin-T~\citep{liu2021swin}, DAT-T~\citep{xia2022vision}, and our ViTAEv2-S. MS denotes that multi-scale inputs are used during testing.}
    \setlength{\tabcolsep}{0.01\linewidth}{\begin{tabular}{c|c|cc|c}
    \hline
    \multirow{2}[2]{*}{{backbone}} & \multirow{2}[2]{*}{{Venue}} & \multirow{2}[2]{*}{{mIoU}} & {mIoU } & {params} \\
          &       &       & {(MS)} & { (M)} \\
    \hline
    ResNet-50 & CVPR'16 & 42.1  & 42.9  & 67 \\
    Swin-T & ICCV'21 & 44.5  & 45.8  & 60 \\
    DAT-T & CVPR'22 & 45.5  & 46.4  & 60 \\
    {ViTAEv2-S} & - & {45.0} & {48.0} & {49} \\
    \hline
    \end{tabular}}%
  \label{tab:segmentationresults}%
\end{table}%

\subsubsection{The performance on semantic segmentation}

\textbf{Settings} We evaluate the ViTAEv2's performance on the semantic segmentation task on the ADE20K~\citep{zhou2017scene,zhou2019semantic} dataset. The ADE20K dataset covers 150 semantic categories with 20K images for training and 2K for validation. We adopt UperNet~\citep{xiao2018unified} as the segmentation framework and train the UperNet with ViTAEv2-S as backbone with default setting used in mmsegmentation~\citep{mmseg2020}, \ie, using the AdamW~\citep{loshchilov2017decoupled} optimizer and fixed image size $512 \times 512$. The models are trained for 160K iterations with a polynomial learning rate decay scheduler. 

\noindent \textbf{Results} The results can be found in Table~\ref{tab:segmentationresults}. With 10M fewer parameters, the segmentation model with ViTAEv2-S as the backbone obtains 45.0 mIoU and outperforms the counterparts using either ResNet or Swin transformer significantly. Besides, when tested with the multi-scale input, the segmentation model with ViTAEv2-S as the backbone obtains much better performance than others, \ie, obtaining 48.0 mIoU. It implies that the ViTAE model can better extract the multi-scale feature owing to the introduced scale-invariance inductive bias. Therefore, it can be benefited more from the multi-scale input.

\subsubsection{The performance on pose estimation}

\begin{table}[htbp]
  \centering
  \setlength{\tabcolsep}{0.01\linewidth}{\caption{Pose estimation results on the AP-10K~\citep{yu2021ap} test set.}
    \begin{tabular}{c|ccc}
    \hline
          & ResNet & Swin-T & {ViTAEv2-S} \\
    \hline
    \#Params & 34.1 M & 32.8 M & {23.1 M} \\
    \hline
    $AP$    & 0.681 & 0.689 & {0.718} \\
    $AP_{50}$ & 0.923 & 0.931 & {0.939} \\
    $AP_{75}$ & 0.718 & 0.751 & {0.786} \\
    \hline
    $AR$    & 0.718 & 0.727 & {0.751} \\
    $AR_{50}$ & 0.933 & 0.939 & {0.947} \\
    $AR_{75}$ & 0.776 & 0.790 & {0.814} \\
    \hline
    \end{tabular}}%
  \label{tab:ViTAEPose}%
\end{table}%

\noindent \textbf{Settings} We evaluate the models' performance on the animal pose estimation task on the AP10K~\citep{yu2021ap} dataset. The AP-10K dataset contains 50 different animal species with animal keypoint annotations. Compared with human pose estimation tasks, animal pose estimation is more challenging due to the diverse species, less labeled data for each species, and significant appearance variance. Therefore, it is more suitable to evaluate the model's generalization ability on this task. Following the setting in AP-10K, we adopt SimpleBaseline~\citep{xiao2018simple} as the pose estimation framework and train the models with various backbones for 210 epochs using Adam optimizer and images of size $256 \times 256$. A step-wise learning rate decay scheduler is employed, and the learning rate is reduced by a factor of 10 after 170 and 200 epochs.

\noindent \textbf{Results} The results are summarized in Table~\ref{tab:ViTAEPose}. As can be seen, the proposed ViTAEv2-S model has fewer parameters yet brings an absolute 3\% AP performance gain over the ResNet-50 backbone. Besides, it also outperforms the Swin-T~\citep{liu2021swin} backbone, especially in the more strict evaluation metric, \ie, $AP_{75}$. These results further demonstrate the superiority of the proposed ViTAEv2 model, which can better handle the tasks with limited data but rich categories, owing to the introduced inductive bias. Recently, it has been shown that the isotropic ViTAE can also achieve superior performance in human pose estimation~\citep{xu2022vitpose}. More research efforts could be made to establish a foundation model for various pose estimation tasks based on ViTAE.

\subsection{Robustness} 
ViTAE employs parallel PCM module and attention module to jointly extract features from both local and global perceptive. Since the two modules extract features in a complementary manner, it is interesting to explore whether such design will make the backbone network robust to adversarial attack~\citep{bhojanapalli2021understanding}. As demonstrated in a recent study~\citep{tang2021robustart} which benchmarks the robustness a series of vision transformer models, CNN models, and MLP models, ViTAE model obtains better robustness under $l_{\infty}$ attack compared with ViT~\citep{dosovitskiy2020image}, MLPMixer~\citep{tolstikhin2021mlp}, and ResNet~\citep{he2016deep}. The theoretical foundation of vision transformer and its variants is expected to be established to explain why introducing the inductive bias into vision transformers can help improve the robustness. 
\section{Discussions}
\label{sec:ViTAElimit}
This paper explores different types of IBs and incorporates them into transformers through the proposed reduction and normal cells. With the collaboration of these two cells, our ViTAE model achieves impressive performance on the ImageNet with fast convergence and high data efficiency. According to the attention distance analysis shown in Figure~\ref{fig:attentionDistance}, the ensemble nature enables the transformer and convolution layers to focus on what they are good at, \ie, modeling long-range dependencies and locality, respectively. As illustrated in Figure~\ref{fig:structure}, our ViTAE model can be viewed as an intra-cell ensemble of complementary transformer layers and convolution layers owing to the skip connection and parallel structure. Moreover, the inductive bias also benefits the transformer models at larger model sizes, \ie, ViTAE-H, or on larger datasets, \ie, ImageNet-22K. Besides, we explore the multi-stage design of ViTAE models and propose ViTAEv2 accordingly, which obtains SOTA performance on image classification and downstream vision tasks, including object detection, semantic segmentation, and pose estimation. More kinds of intrinsic or learnable IBs~\citep{sabour2017dynamic,zhang2022vsa} such as constituting viewpoint invariance can be explored in the future study. On the other hand, although the proposed parallel structure obtains comparable inference speed with better performance, it may also slow down the training depending on the deep learning framework, \eg, dynamic computation graph frameworks like PyTorch need to compute the parallel branches sequentially. Alternatively, static computation graph frameworks like TensorFlow can be adopted to mitigate this issue.

\section{Conclusion}
\label{sec:ViTAEconclud}

In this paper, we incorporate two types of intrinsic inductive bias (IB), i.e., locality and scale-invariance, via reduction and normal cells. By stacking the two cells in both isotropic and multi-stage manner, the proposed ViTAE and ViTAEv2 model obtains superior performance and data efficiency. Specially, extensive experiments show that the multi-stage ViTAEv2 outperforms representative vision transformers in various respects, including classification accuracy, data efficiency, and generalization ability on downstream tasks. When scaling to large-scale models, the inductive bias still helps in improving vision transformers' performance. In future work, we can explore other kinds of IBs to improve their performance further. We hope that this study will provide valuable insights to the following studies introducing intrinsic IB into vision transformers and understanding the impact of intrinsic and learned IBs.

{\small
\bibliographystyle{spbasic}
\bibliography{main}
}

\end{document}